\documentclass[pdflatex,sn-mathphys-ay]{sn-jnl}

\usepackage{graphicx}%
\usepackage{multirow}%
\usepackage{amsfonts}%
\usepackage[title]{appendix}%
\usepackage{xcolor}%
\usepackage{textcomp}%
\usepackage{manyfoot}%
\usepackage{booktabs}%
\usepackage{algorithmicx}%
\usepackage{listings}%

\usepackage{amsmath}    
\usepackage{amssymb}    
\usepackage{amsbsy}     
\usepackage{amsthm}     
\usepackage{mathtools}  
\usepackage{mathrsfs}   
\usepackage{bm}         
\usepackage{eqnarray}   
\usepackage{natbib}     
\usepackage{algorithm}  
\usepackage{algpseudocode}
\usepackage{rotating}   
\usepackage{caption}
\usepackage{adjustbox}

\usepackage{float} 
\usepackage{array}
\usepackage{adjustbox} 
\usepackage{tabularx} 
\usepackage{multirow}
\usepackage{booktabs}
\usepackage{graphicx}
\usepackage{tikz}
\usepackage{forest}

\definecolor{lightblue}{RGB}{199, 222, 255}
\definecolor{lightyellow}{RGB}{255, 236, 179}
\definecolor{lightbrown}{RGB}{215, 204, 200}
\definecolor{lightgreen}{RGB}{200, 230, 201}
\definecolor{lightpurple}{RGB}{225, 190, 231}
\definecolor{lightgrey}{RGB}{230, 230, 230}

\theoremstyle{thmstyleone}%
\theoremstyle{thmstyletwo}%

\theoremstyle{thmstylethree}%

\renewcommand{\hat}{\widehat}

\def \bstheta{\bfsym \theta}

\def\a\cos{\mathrm{arc\cos}}

\newcommand{\bfsym}[1]{\ensuremath{\boldsymbol{#1}}}

\def\bsx{\bfsym x}

\def\boldq{\boldsymbol q}
\def\boldp{\boldsymbol p}

\def\bstheta{\bfsym {\theta}}

\def\bsx{\bfsym x}

\def\bstheta{\bfsym \theta}

\begin{document}

\title[Article Title]{Knowledge Distillation and Dataset Distillation of Large Language Models: Emerging Trends, Challenges, and Future Directions
\footnote{\small{This version of the article has been accepted for publication after peer review but is not the Version of Record.
The Version of Record is available at: \url{https://doi.org/10.1007/s10462-025-11423-3}.}}
}








\author[1]{\fnm{Luyang} \sur{Fang}}
\equalcont{These authors contributed equally to this work.}

\author[2]{\fnm{Xiaowei} \sur{Yu}}
\equalcont{These authors contributed equally to this work.}

\author[1]{\fnm{Jiazhang} \sur{Cai}}
\author[3]{\fnm{Yongkai} \sur{Chen}}
\author[1]{\fnm{Shushan} \sur{Wu}}
\author[4]{\fnm{Zhengliang} \sur{Liu}}
\author[4]{\fnm{Zhenyuan} \sur{Yang}}
\author[1]{\fnm{Haoran} \sur{Lu}}
\author[1]{\fnm{Xilin} \sur{Gong}}
\author[1]{\fnm{Yufang} \sur{Liu}}
\author[5]{\fnm{Terry} \sur{Ma}}
\author[4]{\fnm{Wei} \sur{Ruan}}
\author[6]{\fnm{Ali} \sur{Abbasi}}
\author[2]{\fnm{Jing} \sur{Zhang}}
\author[1]{\fnm{Tao} \sur{Wang}}
\author[7]{\fnm{Ehsan} \sur{Latif}}
\author[4]{\fnm{Weihang} \sur{You}}
\author[4]{\fnm{Hanqi} \sur{Jiang}}
\author[8]{\fnm{Wei} \sur{Liu}}
\author[9]{\fnm{Wei} \sur{Zhang}}
\author[6]{\fnm{Soheil} \sur{Kolouri}}
\author[7]{\fnm{Xiaoming} \sur{Zhai}}
\author[2]{\fnm{Dajiang} \sur{Zhu}}
\author*[1]{\fnm{Wenxuan} \sur{Zhong}}\email{wenxuan@uga.edu}
\author*[4]{\fnm{Tianming} \sur{Liu}}\email{tliu@uga.edu}
\author*[1]{\fnm{Ping} \sur{Ma}}\email{pingma@uga.edu}

\affil*[1]{\orgdiv{Department of Statistics}, \orgname{University of Georgia}, \orgaddress{\city{Athens}, \state{GA}, \country{USA}}}
\affil[2]{\orgdiv{Department of Computer Science and Engineering}, \orgname{The University of Texas at Arlington}, \orgaddress{\city{Arlington}, \state{TX}, \country{USA}}}
\affil[3]{\orgdiv{Department of Statistics}, \orgname{Harvard University}, \orgaddress{\city{Cambridge}, \state{MA}, \country{USA}}}
\affil*[4]{\orgdiv{School of Computing}, \orgname{University of Georgia}, \orgaddress{\city{Athens}, \state{GA}, \country{USA}}}
\affil[5]{\orgdiv{School of Computer Science}, \orgname{Carnegie Mellon University}, \orgaddress{\city{Pittsburgh}, \state{PA}, \country{USA}}}
\affil[6]{\orgdiv{Department of Computer Science}, \orgname{Vanderbilt University}, \orgaddress{\city{Nashville}, \state{TN}, \country{USA}}}
\affil[7]{\orgdiv{AI4STEM Education Center}, \orgname{University of Georgia}, \orgaddress{\city{Athens}, \state{GA}, \country{USA}}}
\affil[8]{\orgdiv{Department of Radiation Oncology}, \orgname{Mayo Clinic Arizona}, \orgaddress{\city{Phoenix}, \state{AZ}, \country{USA}}}
\affil[9]{\orgdiv{School of Computer and Cyber Sciences}, \orgname{Augusta University}, \orgaddress{\city{Augusta}, \state{GA}, \country{USA}}}

\newpage


\abstract{The exponential growth of Large Language Models (LLMs) continues to highlight the need for efficient strategies to meet ever-expanding computational and data demands. This survey provides a comprehensive analysis of two complementary paradigms: Knowledge Distillation (KD) and Dataset Distillation (DD), both aimed at compressing LLMs while preserving their advanced reasoning capabilities and linguistic diversity. We first examine key methodologies in KD, such as task-specific alignment, rationale-based training, and multi-teacher frameworks, alongside DD techniques that synthesize compact, high-impact datasets through optimization-based gradient matching, latent space regularization, and generative synthesis. Building on these foundations, we explore how integrating KD and DD can produce more effective and scalable compression strategies. Together, these approaches address persistent challenges in model scalability, architectural heterogeneity, and the preservation of emergent LLM abilities.
We further highlight applications across domains such as healthcare and education, where distillation enables efficient deployment without sacrificing performance. Despite substantial progress, open challenges remain in preserving emergent reasoning and linguistic diversity, enabling efficient adaptation to continually evolving teacher models and datasets, and establishing comprehensive evaluation protocols. By synthesizing methodological innovations, theoretical foundations, and practical insights, our survey charts a path toward sustainable, resource-efficient LLMs through the tighter integration of KD and DD principles.}

\keywords{Large Language Models, Knowledge Distillation, Dataset Distillation, Efficiency, Model Compression, Survey}



\maketitle

\section{Introduction}\label{sec:intro}

The emergence of Large Language Models (LLMs) like GPT-4 \citep{brown2020language}, DeepSeek \citep{guo2025deepseek}, and LLaMA \citep{touvron2023llama} has transformed natural language processing, enabling unprecedented capabilities in tasks like translation, reasoning, and text generation. Despite these landmark achievements, these advancements come with significant challenges that hinder their practical deployment. First, LLMs demand immense computational resources, often requiring thousands of GPU hours for training and inference, which translates to high energy consumption and environmental costs. Second, their reliance on massive training datasets raises concerns about data efficiency, quality, and sustainability, as public corpora become overutilized and maintaining diverse, high-quality data becomes increasingly difficult \citep{hadi2023survey}. 

To surmount these challenges, distillation has emerged as a pivotal strategy, integrating Knowledge Distillation (KD) \citep{hinton2015distilling} and Dataset Distillation (DD) \citep{wang2018dataset}, to tackle both model compression and data efficiency. Crucially, the success of KD in LLMs hinges on DD techniques, which enable the creation of compact, information-rich synthetic datasets that encapsulate the diverse and complex knowledge of the teacher LLMs.

KD transfers knowledge from a large, pre-trained \textit{teacher} model to a smaller, more efficient \textit{student} model by aligning outputs or intermediate representations. While effective for moderate-scale teacher models, traditional KD struggles with LLMs due to their vast scale, where knowledge is distributed across billions of parameters and intricate attention patterns. Moreover, the knowledge is not limited to output distributions or intermediate representations but also includes higher-order capabilities such as reasoning ability and complex problem-solving skills \citep{wilkins2024higher,zhao2023mskd,latif2024systematic}. DD aims to condense large training datasets into compact synthetic datasets that retain the essential information required to train models efficiently. Recent work has shown that DD can significantly reduce the computational burden of LLM training while maintaining performance. For example, DD can distill millions of training samples into a few hundred synthetic examples that preserve task-specific knowledge \citep{cazenavette2022dataset,maekawa2024dilm}. When applied to LLMs, DD acts as a critical enabler for KD: it identifies high-impact training examples that reflect the teacher’s reasoning processes, thereby guiding the student to learn efficiently without overfitting to redundant data \citep{sorscher2022beyond}.

The scale of LLMs introduces dual challenges: reliance on unsustainable massive datasets \citep{hadi2023survey} and emergent abilities (e.g., chain-of-thought (CoT) reasoning \citep{wei2022emergent}) requiring precise and sophisticated knowledge transfer techniques. These challenges necessitate a dual focus on KD and DD. While KD compresses LLMs by transferring knowledge to smaller models, traditional KD alone cannot address the data efficiency crisis: training newer LLMs on redundant or low-quality data yields diminishing returns \citep{albalak2024survey}. DD complements KD by curating compact, high-fidelity datasets (e.g., rare reasoning patterns \citep{li2024mode}), as demonstrated in LIMA, where 1,000 examples achieved teacher-level performance \citep{zhou2023lima}. This synergy leverages KD’s ability to transfer learned representations and DD’s capacity to generate task-specific synthetic data that mirrors the teacher’s decision boundaries. Together, they address privacy concerns, computational overhead, and data scarcity, enabling smaller models to retain both the efficiency of distillation and the critical capabilities of their larger counterparts.

This survey comprehensively examines KD and DD techniques for LLMs, followed by a discussion of their integration. Traditional KD transfers knowledge from large teacher models to compact students, but modern LLMs’ unprecedented scale introduces challenges like capturing emergent capabilities and preserving embedded knowledge. DD addresses these challenges by synthesizing smaller, high-impact datasets that retain linguistic, semantic, and reasoning diversity for effective training. Our analysis prioritizes standalone advancements in KD and DD while exploring their combined potential to enhance model compression, training efficiency, and resource-aware deployment. This survey underscores their collective role in overcoming scalability, data scarcity, and computational barriers.

While some prior surveys on KD and DD are available, our survey distinguishes itself from them in several significant respects.
This survey, to the best of our knowledge, is the first to place KD and DD under a unifying framework, demonstrating how to jointly utilize them to compact LLMs without losing the capacity for reasoning. Earlier surveys, such as \citet{xu2024survey} and \citet{yu2023dataset}, treat the two subjects as separate, thus overlooking their interplays. We also cover newly developed techniques, such as rationale-based KD, uncertainty-aware KD, and generative model-based DD, which are particularly valuable for the field of LLMs. Moreover, we provide a summary of current theoretical guarantees, converting empirical success to explicit conditions for informing practice and future work. To enrich the DD landscape, we discuss modern data-selection methods that predate LLMs but offer useful ideas for distilled datasets. Additionally, we carefully review evaluation protocols that cover reasoning retention, calibration, robustness, memory usage, and compression level, which provide a useful toolkit for benchmarking distilled LLMs and reveal the compression-performance trade-offs in real-world settings.
By integrating KD and DD, highlighting advanced methods and theories, incorporating data selection insights, and presenting a comprehensive evaluation toolkit, our survey bridges model- and data-centric views, offering a clear blueprint for building reliable and efficient LLMs.

The subsequent sections explore the following key aspects:
\begin{itemize}
    \item Fundamentals of KD and DD (Section~\ref{sec:founda}), distinguishing their roles in compressing LLMs and optimizing training efficiency.
    \item Methodologies for KD in LLMs (Section~\ref{sec:method_KD}), including rationale-based distillation, uncertainty-aware approaches, multi-teacher frameworks, dynamic/adaptive strategies, and task-specific distillation. Additionally, we review theoretical studies that offer deeper insights into the underlying principles of KD.
    \item Methodologies for DD in LLMs (Section~\ref{sec:method_DD}), covering optimization-based distillation, synthetic data generation, and complementary data selection strategies for compact training data.
    \item Integration of KD and DD (Section~\ref{sec:combine}), presenting unified frameworks that combine KD and DD strategies for enhancing LLMs.
    \item Evaluation metrics (Section~\ref{sec:eval}) for assessing the effectiveness of distillation in LLMs, focusing on performance retention, computational efficiency, and robustness.
    \item Applications across multiple domains (Section~\ref{sec:apply}), including medical and health, education, and bioinformatics, demonstrating the practical benefits of distillation in real-world scenarios.
    \item Challenges and future directions (Section~\ref{sec:challenge}), identifying key areas for improvement.
\end{itemize}

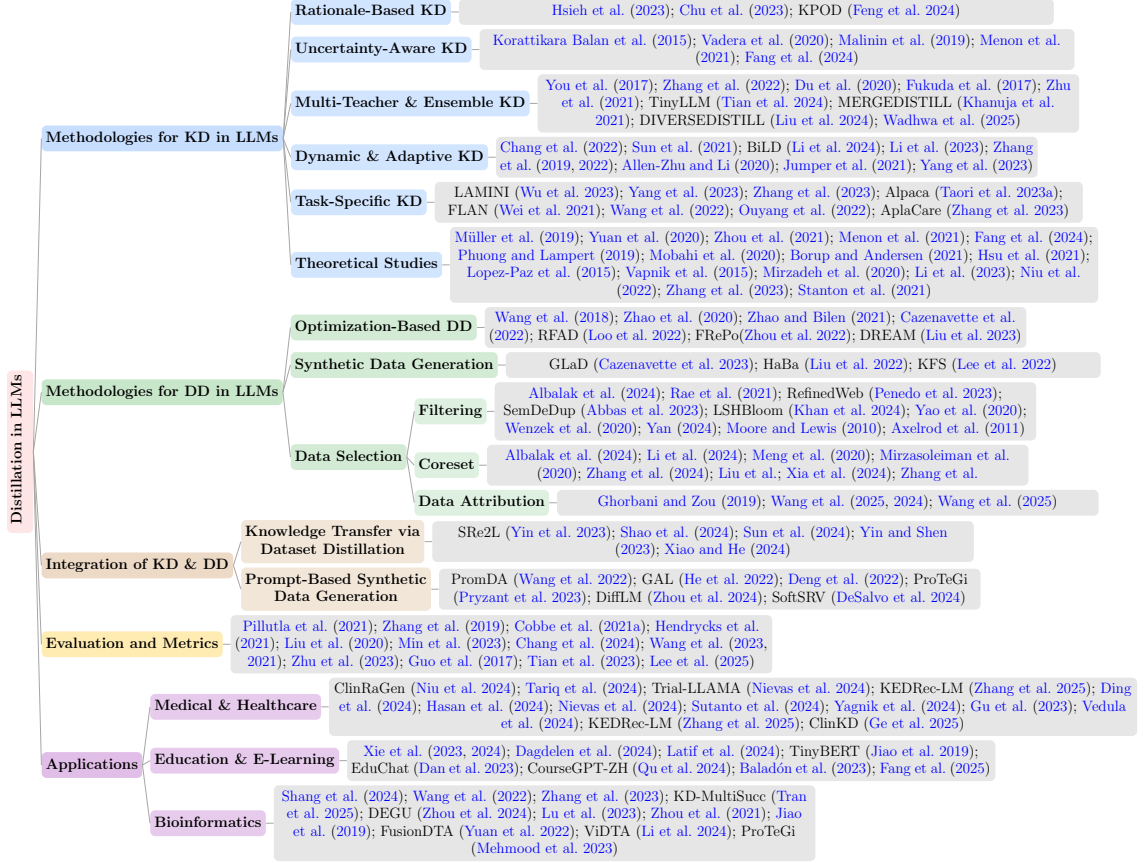
\begin{figure}[h]
\centering
\resizebox{1.0\textwidth}{!}{
\begin{forest}
for tree={
    grow=east,
    parent anchor=east,
    child anchor=west,
    edge={draw=gray!70, thick},
    align=center,
    font=\large,
    anchor=west,
    l sep=6pt,
    s sep=3pt,
    before computing xy={sibling reversed},
    rounded corners
}
[ \rotatebox{90}{\textbf{Distillation in LLMs}}, fill=red!10, rounded corners,
    text width=0.5cm,
  [\textbf{Applications}, fill=lightpurple
    [\textbf{Bioinformatics}, fill=lightpurple!80
      [
        \parbox{1.0\linewidth}{\centering
         \citet{shang2024accurate,wang2022contact,zhang2023adaptive};
         KD-MultiSucc \citep{tran2025kd_multisucc};
         DEGU \citep{zhou2024uncertainty};
         \citet{lu2023improving,zhou2021improving,jiao2019tinybert};
         FusionDTA \citep{yuan2022fusiondta};
         ViDTA \citep{li2024vidta};
         ProTeGi \citep{mehmood2023distilling}
        },
        fill=lightgrey
      ]
    ]
    [\textbf{Education \& E-Learning}, fill=lightpurple!80
      [
        \parbox{1.2\linewidth}{\centering
          \citet{xie2023darwin,xie2024darwin,dagdelen2024structured,latif2024knowledge};
          TinyBERT \citep{jiao2019tinybert};
          EduChat \citep{dan2023educhat};
          CourseGPT-ZH \citep{qu2024coursegpt};
          \citet{baladon2023retuyt,fang2025efficient}
        },
        fill=lightgrey
      ]
    ]
    [\textbf{Medical \& Healthcare}, fill=lightpurple!80
      [
        \parbox{1.5\linewidth}{\centering
         ClinRaGen \citep{niu2024clinragen};
         \citet{tariq2024radiology};
         Trial-LLAMA \citep{nievas2024clinicaltrials};
         KEDRec-LM \citep{zhang2025kedrec};
         \citet{ding2024ckle,hasan2024optimclm,nievas2024clinicaltrials,sutanto2024llmqa,yagnik2024medlm,gu2023ade,vedula2024cie};
         KEDRec-LM \citep{zhang2025kedrec};
         ClinKD \citep{ge2025clinkd}
        },
        fill=lightgrey
      ]
    ]
  ]
  [\textbf{Evaluation and Metrics}, fill=lightyellow
    [
      \parbox{1.0\linewidth}{\centering
        \citet{pillutla2021mauve,zhang2019bertscore,cobbe2021training,hendrycks2021measuring,liu2020logiqa};
        \citet{min2023factscore,chang2024survey,wang2023robustness,wang2021adversarial,zhu2023promptbench};
        \citet{guo2017calibration,tian2023just};
        \citet{lee2025distillation}
      },
      fill=lightgrey
    ]
  ]
  [\textbf{Integration of KD \& DD} , fill=brown!30
     [\textbf{Prompt-Based Synthetic}\\ \textbf{Data Generation}, fill=brown!20
        [
          \parbox{1.0\linewidth}{\centering
            PromDA \citep{wang2022promda}; 
            GAL \citep{he2022generate};
            \citet{deng2022rlprompt};
            ProTeGi \citep{pryzant2023automatic}; 
            DiffLM \citep{zhou2024difflm};
            SoftSRV \citep{desalvo2024no}
          },
          fill=lightgrey
        ]
      ]
      [\textbf{Knowledge Transfer via} \\ \textbf{Dataset Distillation}, fill=brown!20
        [
          \parbox{1.0\linewidth}{\centering
            SRe2L \citep{yin2023squeeze};
            \citet{shao2024generalized,sun2024diversity,yin2023dataset,xiao2024large}
          },
          fill=lightgrey
        ]
      ]
  ]
  [\textbf{Methodologies for DD in LLMs}, fill=lightgreen
    [\textbf{Data Selection}, fill=lightgreen!80
      [\textbf{Data Attribution}, fill=lightgreen!60
        [
          \parbox{1.0\linewidth}{\centering
            \citet{ghorbani2019data,wang2024data,wangrethinking}; 
            \citet{wang2024capturing}
          },
          fill=lightgrey
        ]
      ]
      [\textbf{Coreset}, fill=lightgreen!60
        [
          \parbox{1.0\linewidth}{\centering
            \citet{albalak2024survey,li2024core,meng2020more,mirzasoleiman2020coresets,zhang2024tagcos,liumakes,xia2024less,zhangstaff}
          },
          fill=lightgrey
        ]
      ]
      [\textbf{Filtering}, fill=lightgreen!60
        [
          \parbox{1.0\linewidth}{\centering
           \citet{albalak2024survey,rae2021scaling};
           RefinedWeb \citep{penedo2023refinedweb};
           SemDeDup \citep{abbas2023semdedup};
           LSHBloom \citep{khan2024lshbloom};
           \citet{yao2020text};
           \citet{wenzek2020ccnet,yan2024optimizing,moore2010intelligent, axelrod2011domain}
          },
          fill=lightgrey
        ]
      ]
    ]
    [\textbf{Synthetic Data Generation}, fill=lightgreen!80
      [
        \parbox{1.1\linewidth}{\centering
         GLaD \citep{cazenavette2023generalizing};
         HaBa \citep{liu2022dataset};
         KFS \citep{lee2022dataset}
        },
        fill=lightgrey
      ]
    ]
    [\textbf{Optimization-Based DD}, fill=lightgreen!80
      [
        \parbox{1.0\linewidth}{\centering
         \citet{wang2018dataset,zhao2020dataset}; 
         \citet{zhao2021dataset,cazenavette2022dataset};
         RFAD \citep{loo2022efficient};
         FRePo\citep{zhou2022dataset};
         DREAM \citep{liu2023dream}
        },
        fill=lightgrey
      ]
    ]
  ]
  [\textbf{Methodologies for KD in LLMs}, fill=lightblue
    [\textbf{Theoretical Studies}, fill=lightblue!80
      [
        \parbox{1.2\linewidth}{\centering
        \citet{muller2019does,yuan2020revisiting,zhou2021rethinking,menon2021statistical,fangbayesian2024,phuong2019towards,mobahi2020self,borup2021even,hsu2021generalization,lopez2015unifying,vapnik2015learning,mirzadeh2020improved,li2023kd,niu2022respecting,zhang2023towards,stanton2021does}
        },
        fill=lightgrey
      ]
    ]
    [\textbf{Task-Specific KD}, fill=lightblue!80
      [
        \parbox{1.2\linewidth}{\centering
         LAMINI \citep{wu2023lamini};
         \citet{yang2023enabling};
         \citet{zhang2023instruction};
         Alpaca \citep{taori2023alpaca};
         FLAN \citep{wei2021finetuned};
         \citet{wang2022self,ouyang2022training};
         AplaCare \citep{zhang2023alpacare}
        },
        fill=lightgrey
      ]
    ]
    [\textbf{Dynamic \& Adaptive KD}, fill=lightblue!80
      [
        \parbox{1.0\linewidth}{\centering
         \citet{chang-etal-2022-one}; 
         \citet{sun2021collaborative}; 
         BiLD \citep{li2024bild};
         \citet{li2023unlock,Zhang2019BeYO,Zhang2022SelfDistillation,DBLP:journals/corr/abs-2012-09816,jumper2021highly,yang2023alphafold2}
        },
        fill=lightgrey
      ]
    ]
    [\textbf{Multi-Teacher \& Ensemble KD}, fill=lightblue!80
      [
        \parbox{1.0\linewidth}{\centering
         \citet{you2017learning};
         \citet{zhang2022confidence}; 
         \citet{du2020agree};
         \citet{fukuda2017efficient}; 
         \citet{zhu2021data};
         TinyLLM \citep{tian2024beyond};
         MERGEDISTILL \citep{khanuja2021mergedistill};
         DIVERSEDISTILL \citep{liu2024wisdom};
         \citet{wadhwa2025taught}
        },
        fill=lightgrey
      ]
    ]
    [\textbf{Uncertainty-Aware KD}, fill=lightblue!80
      [
        \parbox{1.1\linewidth}{\centering
         \citet{korattikara2015bayesian};
         \citet{vadera2020generalized};
         \citet{malinin2019ensemble};
         \citet{menon2021statistical};
         \citet{fangbayesian2024}
        },
        fill=lightgrey
      ]
    ]
    [\textbf{Rationale-Based KD}, fill=lightblue!80
      [
        \parbox{1.1\linewidth}{\centering
         \citet{hsieh2023distilling}; 
         \citet{chu2023survey}; 
         KPOD \citep{feng2024keypoint}
        },
        fill=lightgrey
      ]
    ]
  ]
]
\end{forest}}
\caption{Taxonomy of Distillation of Large Language Models.}
\label{fig:taxonomy}
\end{figure}

The taxonomy of this survey is illustrated in Figure~\ref{fig:taxonomy}.

\section{Fundamentals of Distillation}\label{sec:founda}

This section introduces the definition and core concepts of Knowledge Distillation (KD) and Dataset Distillation (DD). Table~\ref{KDvsDD} presents a comparative summary of KD and DD. Knowledge distillation has consistently demonstrated high effectiveness across a wide range of benchmarks such as GLUE, SuperGLUE, and MMLU, where student models often retain over 95\% of the teacher model’s performance while offering significant efficiency gains. These results establish KD as a reliable approach for compressing large models without substantial loss in accuracy. In contrast, dataset distillation is a more recent technique, with performance evaluations on datasets like ImageNet, AlpacaEval, and GSM8K showing promising results—often achieving 80–90\% of the performance obtained using full real datasets. However, DD’s scalability to large-scale language models remains under active investigation. While KD is widely adopted in production for its stability and maturity, DD is gaining traction for scenarios requiring data efficiency, privacy preservation, or decentralized training environments. The following subsections discuss the significance of distillation in LLMs compared to traditional distillation methods.

\begin{table*}[htbp]
\centering
\caption{Comparison of Knowledge Distillation and Dataset Distillation.}\label{KDvsDD}
\resizebox{0.95\textwidth}{!}{
\begin{tabular}{p{3.5cm}p{6.25cm}p{6.25cm}}
\toprule
\textbf{Aspect} & \textbf{Knowledge Distillation (KD)} & \textbf{Dataset Distillation (DD)} \\
\midrule
Goal & Transfer knowledge from a large teacher model to a smaller student & Synthesize a small dataset that approximates the training signal \\
\midrule
Output & A compact student model & A small, synthetic dataset \\
\midrule
Efficiency Gains & Faster inference, reduced deployment cost & Faster training, reduced data storage or communication costs \\
\midrule
Data Requirements & Often requires access to original or augmented data with teacher outputs & Can work with minimal or no access to original data \\
\midrule
Scalability & Scales well to various downstream tasks and modalities & Scalability to large LLMs still limited and under investigation \\
\midrule
Typical Benchmarks & GLUE, SQuAD, MMLU, SuperGLUE, GPQA & ImageNet, AlpacaEval, MT-Bench, MeetingBank, ZeroScrolls, GSM8K \\
\midrule
Performance Trends & Distilled models can retain 95\% of teacher performance & Synthesized data can match $\sim$80--90\% of full data performance on small tasks \\
\midrule
Use Cases & Model compression, efficient inference, domain adaptation & Data-efficient training, privacy-aware learning, federated learning \\
\midrule
Challenges & Maintaining generalization and robustness of student & Quality of synthetic data, difficulty scaling to high-capacity models \\
\bottomrule
\end{tabular}
}
\end{table*}

\subsection{Knowledge Distillation}

\subsubsection{Definition and Core Concepts} \label{sec:Def}

KD is a model compression paradigm that transfers knowledge from a computationally intensive teacher model $f_T$ to a compact student model $f_S$. Formally, KD trains $f_S$ to approximate both the output behavior and intermediate representations of $f_T$. The foundational work of \citet{hinton2015distilling} introduced the concept of soft labels: instead of training on hard labels $y$, the student learns from the teacher's class probability distribution $\boldsymbol{p}_T = \sigma\bigl(\mathbf{z}_T / \tau\bigr)$, where $\mathbf{z}_T$ are logits from $f_T$, $\sigma$ is the softmax function, and $\tau$ is a temperature parameter that controls distribution smoothness. The student's objective combines a cross-entropy loss $\mathcal{L}_{\mathrm{CE}}$ (for hard labels) and a distillation loss $\mathcal{L}_{\mathrm{KL}}$:
\begin{equation}\label{eq: KD loss}
    \mathcal{L}_{\text {KD}} \;=\; \alpha \cdot \mathcal{L}_{\mathrm{CE}}\bigl(\sigma(\mathbf{z}_S(\boldsymbol{x})),\, y\bigr)
    \;+\; (1-\alpha) \cdot \tau^2 \cdot \mathcal{L}_{\mathrm{KL}}\bigl(\sigma(\boldsymbol{z}_T(\boldsymbol{x})/\tau),\, \sigma(\boldsymbol{z}_S(\boldsymbol{x})/\tau)\bigr),
\end{equation}
where $\mathcal{L}_{\mathrm{KL}}$ is the Kullback-Leibler (KL) divergence between student and teacher softened outputs, and $\alpha$ balances the two terms. Beyond logits, later works generalized KD to transfer hidden state activations \citep{romero2014fitnets}, intermediate layers \citep{sun2019patient}, attention matrices \citep{jiao2019tinybert}, or relational knowledge \citep{park2019relational}, formalized as minimizing distance metrics (e.g., $\|\boldsymbol{h}_T-\boldsymbol{h}_S\|^2$) between teacher and student representations. This framework enables the student to inherit not only task-specific accuracy but also the teacher's generalization patterns, making KD a cornerstone for efficient model deployment.

\begin{figure}[h]
\centering
\includegraphics[width=0.9\linewidth]{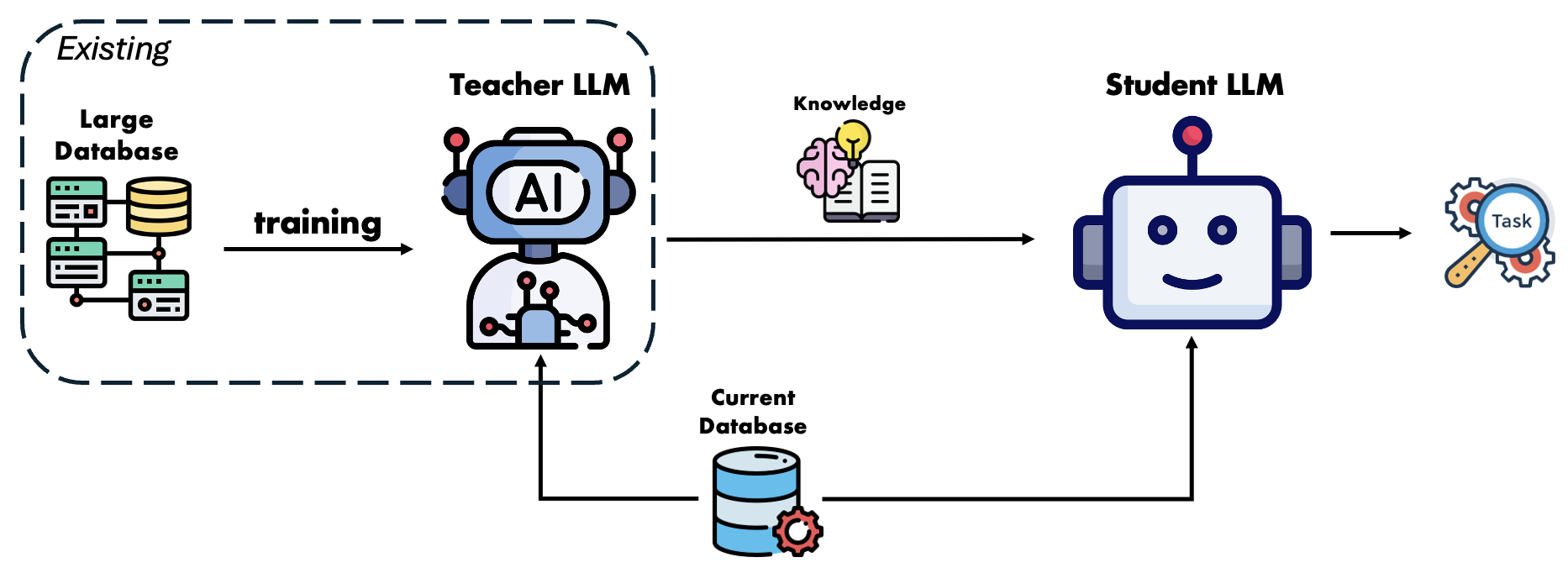}
\caption{Overview of Knowledge Distillation in LLMs. Knowledge is distilled from a teacher LLM, which has been trained on a large existing database. This knowledge, potentially enriched with current, task-specific data, is transferred to a smaller student LLM. By learning from both the teacher’s guidance and the current data, the student LLM becomes more efficient and effective at performing downstream tasks.}
\label{fig:KD_frame_LLM}
\end{figure}

\subsubsection{KD in the Era of LLMs vs. Traditional Models}

The emergence of LLMs, exemplified by models like GPT-3 (175B parameters \citep{brown2020language}), has necessitated rethinking traditional KD paradigms. While classical KD usually focuses on compressing task-specific models (e.g., ResNet-50 to MobileNet) with homogeneous architectures \citep{gou2021knowledge}, LLM-driven distillation confronts four fundamental shifts:

\begin{itemize}
    \item \textbf{Scale-Driven Shifts:} Traditional KD operates on static output distributions (e.g., class probabilities), but autoregressive LLMs generate sequential token distributions over vocabularies of $\sim$50k tokens. This demands novel divergence measures for sequence-level knowledge transfer \citep{shridhar2023distilling}, such as token-level Kullback-Leibler minimization or dynamic temperature scaling.
    
    \item \textbf{Architectural Heterogeneity:}  Traditional KD often assumed matched or closely related teacher-student topologies (e.g., both CNNs). LLM distillation often bridges architecturally distinct models (e.g., sparse Mixture-of-Experts teachers to dense students \citep{fedus2022switch}). This requires layer remapping strategies \citep{jiao2019tinybert} and representation alignment (e.g., attention head distillation \citep{michel2019sixteen}) to bridge topological gaps while preserving generative coherence.
    
    \item \textbf{Knowledge Localization:} LLMs encode knowledge across deep layer stacks and multi-head attention mechanisms, necessitating distillation strategies that address:
    \begin{itemize}
        \item \textit{Structural patterns:} Attention head significance \citep{michel2019sixteen} and layer-specific functional roles (e.g., syntax vs.\ semantics).
        \item \textit{Reasoning trajectories:} Explicit rationales like CoT and implicit latent state progressions.
    \end{itemize}
    Unlike traditional model distillation, which often focuses on replicating localized features, LLM distillation must preserve cross-layer dependencies that encode linguistic coherence and logical inference \citep{sun2019patient}.
    
    \item \textbf{Dynamic Adaptation:} LLM distillation increasingly employs iterative protocols where teachers evolve via reinforcement learning from human feedback (RLHF) \citep{ouyang2022training} or synthetic data augmentation \citep{taori2023stanford}, diverging from static teacher assumptions in classical KD.
\end{itemize}

\subsection{Dataset Distillation}\label{sec:concept_DD}

\subsubsection{Overview of Dataset Distillation}

Dataset distillation \citep{wang2018dataset} is a technique designed to condense knowledge from large datasets into significantly smaller, synthetic datasets while retaining the ability to train models effectively. Unlike data selection methods (e.g., data pruning or coreset selection \citep{dasgupta2009sampling}), which focus on choosing representative real samples, dataset distillation actively synthesizes new, compact samples that encapsulate the essential learning signal. The distilled dataset is often orders of magnitude smaller yet enables models to achieve comparable or even improved performance.

Formally, let $\mathcal{D} \triangleq \{(\boldsymbol{x}_{i}, y_{i})\}_{i=1}^{|\mathcal{D}|}$ be a large dataset, and $\mathcal{D}_{\text{syn}} \triangleq \{(\tilde{\boldsymbol{x}}_{i}, \tilde{y}_{i})\}_{i=1}^{n}$ be the distilled dataset with $n \ll |\mathcal{D}|$. For a learning model $\Phi$, let $\theta^{\mathcal{D}}$ and $\theta^{\mathcal{D}_{\text{syn}}}$ be the parameters learned from training on $\mathcal{D}$ and $\mathcal{D}_{\text{syn}}$, respectively. Dataset distillation aims to make $\theta^{\mathcal{D}}$ and $\theta^{\mathcal{D}_{\text{syn}}}$ produce similar outcomes:
\begin{equation}
\underset{\mathcal{D}_{\text{syn}}, \, n}{\arg\min}\,
\Bigl(\sup_{\boldsymbol{x} \sim \mathcal{X},\, y \sim \mathcal{Y}}\bigl\{\bigl|\,
l\bigl(\Phi_{\theta^{\mathcal{D}}} (\boldsymbol{x}), y\bigr)
\;-\;
l\bigl(\Phi_{\theta^{\mathcal{D}_{\text{syn}}}} (\boldsymbol{x}), y\bigr)\bigr|\bigr\}\Bigr).
\end{equation}
An $\epsilon$-approximate data summary satisfies:
\begin{equation}
\sup_{\boldsymbol{x} \sim \mathcal{X},\, y \sim \mathcal{Y}}\bigl\{\bigl|\,
l\bigl(\Phi_{\theta^{\mathcal{D}}} (\boldsymbol{x}), y\bigr)
\;-\;
l\bigl(\Phi_{\theta^{\mathcal{D}_{\text{syn}}}} (\boldsymbol{x}), y\bigr)\bigr|\bigr\}
\;\le\;\epsilon.
\end{equation}

\begin{figure}[h]
\centering
\includegraphics[width=0.8\linewidth]{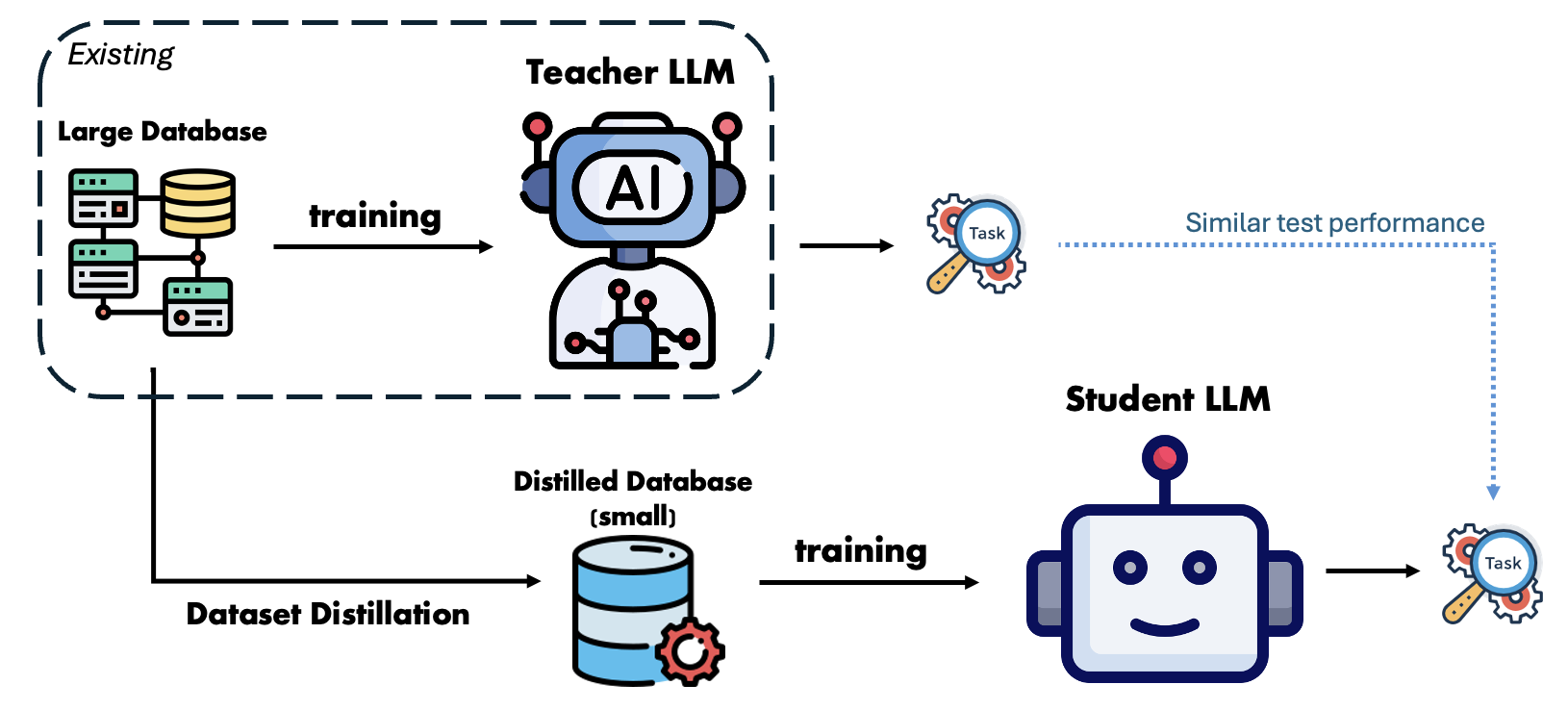}
\caption{Overview of Dataset Distillation in LLMs. A teacher LLM is trained on a massive original database. Through dataset distillation, a compact, high-quality subset (Distilled Database) is synthesized to preserve essential knowledge. This smaller dataset is then used to train a student LLM, aiming to achieve similar performance as the teacher while requiring significantly fewer data.}
\label{fig:DD_frame_LLM}
\end{figure}

With the increasing scale of modern deep learning models, dataset distillation has gained attention for accelerating training \citep{ding2024calibrated}, enabling continual learning \citep{deng2022remember}, and improving data efficiency in low-resource settings \citep{song2023federated}. However, challenges remain, such as preserving sufficient diversity and robustness in the distilled data and avoiding overfitting \citep{cazenavette2023generalizing}. In LLMs, dataset distillation is crucial for reducing computational overhead while maintaining the rich semantic diversity needed for effective language modeling. Table \ref{commondatasets} outlines some commonly used datasets for data distillation.

\begin{table}[htbp]
\centering
\caption{Common Datasets for Data Distillation.}
\renewcommand{\arraystretch}{1}
\begin{tabular}{l c c l}
\toprule
\textbf{Dataset} & \textbf{Size} & \textbf{Category} & \textbf{Related Works} \\
\midrule
SVHN            & 60K   & Image & HaBa~\citep{liu2022dataset}, FreD~\citep{Shin2023FreD} \\
CIFAR-10        & 60K   & Image & FreD~\citep{Shin2023FreD}, IDC~\citep{Kim2022IDC} \\
CIFAR-100       & 60K   & Image & IDM~\citep{zhao2023improved}, DD~\citep{wang2018dataset} \\
TinyImageNet    & 100K  & Image & CMI~\citep{Zhong2024CMI}, DD~\citep{wang2018dataset} \\
ImageNet-1K     & 14000K& Image & Teddy~\citep{Yu2024Teddy}, TESLA~\citep{Cui2023tesla} \\
TDBench         & 23 datasets & Table & TDColER~\citep{Kang2025TDColER} \\
Speech Commands & 8K   & Audio & IDC~\citep{Kim2022IDC} \\
AMC AIME STaR   & 4K   & Text  & Numinamath~\citep{li2024Numinamath} \\
R1-Distill-SFT  & 17K  & Text  & QWTHN~\citep{Kong2025QWTHN} \\
OpenThoughts-114k & 114K & Text  & FreeEvalLM~\citep{zhao2025Trade} \\
\bottomrule
\end{tabular}
\label{commondatasets}
\end{table}

\subsubsection{Methods and Approaches}

Dataset distillation has evolved through various methodological advancements, each aiming to compress large datasets into small yet highly informative subsets \citep{Noveen2023tmlr, Yin2024dataset, yu2023dataset}. From a high-level perspective, two main categories can be distinguished: \textit{optimization-based dataset distillation} and \textit{synthetic data generation for dataset distillation}.

\paragraph{Optimization-Based Methods.} These methods distill datasets by aligning training dynamics (e.g., gradients, trajectories) or final model performance between synthetic and original data.

\begin{itemize}
    \item \textbf{Meta-Learning Optimization:} A bi-level optimization approach that trains a model on a synthetic dataset and evaluates its performance on the original dataset. The synthetic samples are iteratively refined to match the model’s performance when trained on the full dataset \citep{wang2018dataset}.
    \item \textbf{Gradient Matching:} Proposed by \citet{zhao2020dataset}, it aligns the gradients from one-step updates on real vs.\ synthetic datasets. The objective is to make gradients consistent so that training on synthetic data closely resembles training on real data over short time spans.
    \item \textbf{Trajectory Matching:} To address the limitation of single-step gradient matching, multi-step parameter matching (a.k.a.\ trajectory matching) was introduced by \citet{cazenavette2022dataset}. It aligns the endpoints of training trajectories over multiple steps, making the distilled dataset more robust over longer training horizons.
\end{itemize}

\paragraph{Synthetic Data Generation.} These methods directly create artificial data that approximates the distribution of the original dataset. The representative technique here is \textit{Distribution Matching} (DM) \citep{zhao2023dataset}, aiming to generate synthetic data whose empirical distribution aligns with the real dataset using metrics like Maximum Mean Discrepancy (MMD).

\subsubsection{Applications and Use Cases}

Dataset distillation has wide-ranging applications where data redundancy must be reduced without sacrificing performance:

\begin{itemize}
    \item \textbf{LLM Fine-Tuning and Adaptation:} By creating a distilled subset of domain-specific data, researchers can rapidly adapt large-scale LLMs to specialized tasks without incurring the full computational cost.
    \item \textbf{Low-Resource Learning:} In federated learning and edge AI scenarios \citep{Wu2024LLMFL, Qu2025LLMEdge}, a distilled dataset reduces communication and computational overhead.
    \item \textbf{Neural Architecture Search and Hyperparameter Tuning:} Distilled datasets provide a proxy for evaluating model variants quickly, cutting down on expensive full-dataset training \citep{Prabhakar2022DDNAS}.
    \item \textbf{Privacy and Security:} Distilled datasets can serve as privacy-preserving proxies for sensitive data (e.g., medical or financial records), reducing exposure of individual-level information \citep{yu2023dataset}.
    \item \textbf{Continual Learning:} By summarizing past tasks, distilled datasets help mitigate catastrophic forgetting when models learn incrementally over time \citep{Binici2022WACV}.
\end{itemize}

\section{Methodologies and Techniques for Knowledge Distillation in LLMs}\label{sec:method_KD}

In this section, we review methodologies for KD in LLMs, including rationale-based approaches, uncertainty-aware techniques, multi-teacher frameworks, dynamic/adaptive strategies, and task-specific distillation. We also explore theoretical studies that uncover the foundational mechanisms driving KD’s success in LLMs.

\subsection{Rationale-Based KD }\label{Wei Ruan}
Rationale-Based Knowledge Distillation (RBKD) improves traditional knowledge distillation by allowing the student model to learn not only the teacher’s final predictions but also the reasoning process behind them, like CoT reasoning. This makes the student model more interpretable and reduces the need for large amounts of labeled data~\citep{hsieh2023distilling}. Instead of merely imitating the teacher’s outputs, the student develops a deeper understanding of problem-solving, leading to better generalization and adaptability to new tasks.

Formally, given a dataset \(\mathcal{D} = \{(x_i, y_i, r_i)\}_{i=1}^n\), where \(x_i\) is the input, \(y_i\) is the label, and \(r_i\) is the rationale generated by the teacher, the student is trained to jointly predict both the rationale and the final answer. This objective can be formulated as:

\begin{equation}
\mathcal{L}_{\text{RBKD}} = -\frac{1}{n} \sum_{i=1}^n \log P_\theta(r_i, y_i \mid x_i),
\label{eq:rbkd_loss}
\end{equation}
where \(P_\theta\) denotes the student model parameterized by \(\theta\). This formulation encourages the student to internalize the reasoning path rather than shortcutting to the answer.

By incorporating reasoning steps, RBKD enhances transparency, making models more reliable in fields like healthcare and law, where understanding decisions is crucial. It also improves efficiency, as smaller models can achieve strong performance without requiring extensive computational resources. This makes RBKD a practical approach that balances accuracy, interpretability, and resource efficiency~\citep{chu2023survey}.

One promising direction is Keypoint-based Progressive CoT Distillation (KPOD), which addresses both token significance and the order of learning~\citep{feng2024keypoint}. In KPOD, the difficulty of each reasoning step is quantified using a weighted token generation loss:

\begin{equation}
d_k^{(i)} = -\sum_{j=p_k}^{q_k} \hat{w}_j^{(i)} \log P(r_j^{(i)} \mid r_{<j}^{(i)}, x^{(i)}; \theta_s),
\label{eq:step_difficulty}
\end{equation}
where \(d_k^{(i)}\) is the difficulty score of the \(k\)-th step in the \(i\)-th rationale, \(p_k\) and \(q_k\) denote the start and end positions of that step, and \(\hat{w}_j^{(i)}\) represents the normalized importance weight for token \(j\). This difficulty-aware design allows the student model to acquire reasoning skills progressively, from simpler to more complex steps, resulting in stronger generalization and interoperability.

\paragraph{Challenges and Future Directions.}

The growing focus on distilling richer knowledge from teacher LLMs, such as reasoning and other higher-order capabilities, raises important questions about which knowledge to extract and how to extract it effectively. Since many teacher LLMs are closed-source, capturing their advanced capabilities is more challenging than merely collecting hard-label predictions. 
This highlights several specific research gaps that warrant investigation. First, how can we develop standardized metrics to evaluate the quality of extracted reasoning traces, particularly when ground truth reasoning paths are unavailable? 
Second, what novel prompting strategies can reliably elicit intermediate reasoning steps from black-box teacher LLMs without access to their internal representations? 
Third, existing methods primarily focus on single-step reasoning extraction. It is an interesting direction to explore how we can effectively capture and transfer sequential reasoning chains and complex problem decomposition strategies.

\subsection{Uncertainty-Aware KD}\label{Yongkai Chen}
Traditional KD methods mainly focus on matching the predictions or latent representations of the teacher model to improve the generalization of the student model. While effective, this approach overlooks the inherent uncertainty in the teacher’s predictions, which can provide critical insights into noisy samples.
To address this limitation, the study of uncertainty in KD has become an active field driven by two primary goals: (1) distilling the uncertainty from probabilistic teachers and (2) quantifying the uncertainty of the distilled models.

In the first scenario, the teacher model generates a predictive distribution, such as the conditional probability measure $p_T(y|\bsx)$.  When the teacher model is trained as a Bayesian neural network (BNN) \citep{mackay1992practical} or a Monte Carlo ensembles of models, 
$p_T(y|\bsx)$  usually represents the empirical distribution derived from Monte Carlo samples.
Compared to the poor approximation of variational inference in general neural networks or the large computational complexity of the Monte Carlo approximation, Knowledge distillation provides a general and simple way to train a small model $p_S(y|\bsx)$ that provides direct estimation of conditional probability $p_T(y|\bsx)$. This approach usually minimizes the KL divergence between $p_S(y|\bsx)$ and  $p_T(y|\bsx)$.
\citet{korattikara2015bayesian} introduces Bayesian Dark Knowledge (BDK) by parameterizing $p_S(y|\bsx)$ as $[p_S(y = 1|\bsx),\ldots,p_S(y = K|\bsx)]$ for $K$-class classification, and as $\mathcal{N}(y|\mu(\bsx), \sigma^2(\bsx))$ for regression, where $\mathcal{N}$ is the probability density function of the normal distribution with mean $\mu(\bsx)$ and variance $\sigma^2(\bsx)$. The effectiveness is demonstrated in distilling ensembles of teacher models trained using stochastic Langevin Monte Carlo.
Subsequent extensions include \citet{vadera2020generalized}, which generalizes BDK by distilling posterior expectations, and \citet{malinin2019ensemble}, which proposes Ensemble Distribution Distillation to capture both the ensemble mean and diversity within a single model.

\begin{figure}[h]
    \centering
    \includegraphics[width=0.85\linewidth]{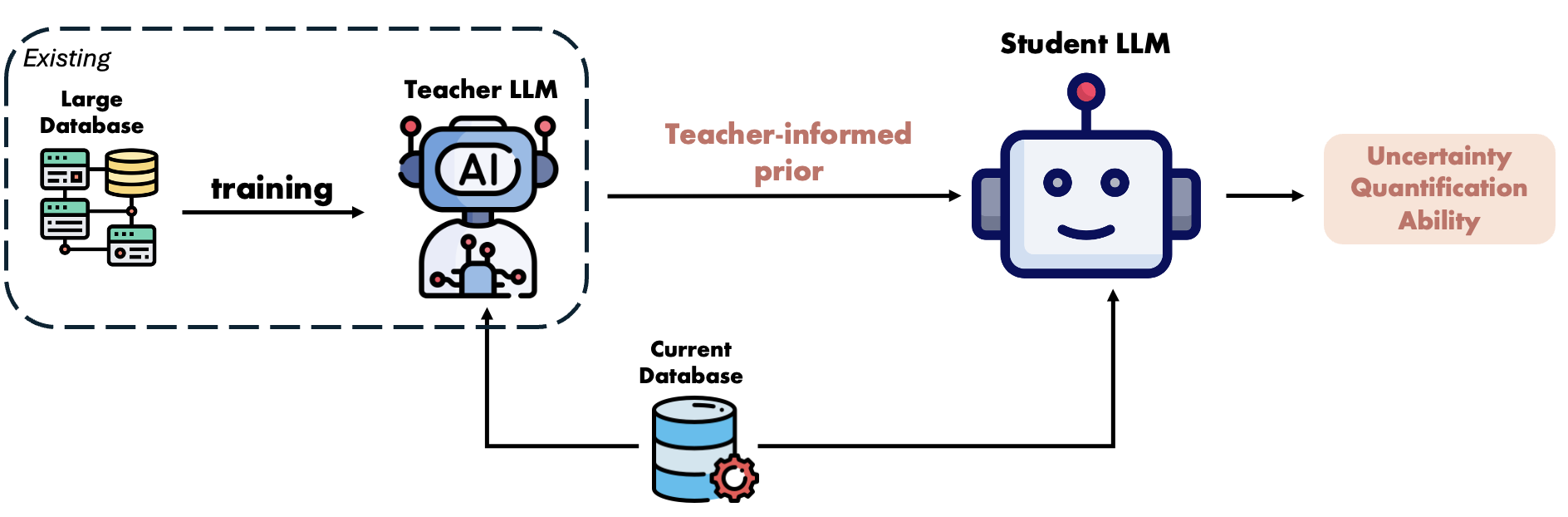}
    \caption{Bayesian Knowledge Distillation. Knowledge distilled from the teacher LLM is incorporated as a teacher-informed prior over the student LLM’s parameters. This leads to a posterior distribution that equips the student LLM with uncertainty quantification capabilities.}
    \label{fig:BKD}
\end{figure}

In the second scenario, the researchers aim to quantify the uncertainty of the distilled models, avoiding the propagation of incorrect knowledge to the student and reducing the risk of making overconfident predictions. A consensus strategy is to interpret KD within a Bayesian inference framework. Specifically, \citet{fangbayesian2024} proposes a systematic statistical Bayesian framework to interpret the distillation process, with the workflow available in Figure \ref{fig:BKD}. Bayesian Knowledge Distillation (BKD) is introduced to establish the equivalence between KD and a Bayesian model.
It defines a proper prior distribution for the student model's parameter based on the prediction outcomes of the teacher models, referred to as the teacher-informed prior (TIP). For example, in the $K$-class classification task, given the predicted class probabilities $\{\boldp_{i}\}_{i=1}^N$, where $\boldp_i=(p_{i1},\ldots p_{iK})^T$, by the teacher model $M_t$, the prior distribution $\pi_{\bstheta}(\bstheta;\{\boldp_{i}\}_{i=1}^N)$ on the student model's weights $\bstheta$ is defined by,
\begin{equation}\label{eq:prior_KD_theta}
    \pi_{\bstheta}(\bstheta;\{\boldp_{i}\}_{i=1}^N) \propto \Pi_{i=1}^N \pi_{\boldq}(\boldq_{i};\boldp_{i}),
\end{equation}
where $\propto$ denotes proportionality, and $\pi_{\boldq}(\boldq;\boldp)$ is a probability density function defined for a $K$-dimensional probability vector $\boldq = (q_1,\ldots,q_K)^T$ with the parameter $\boldp$.
Following the idea of KD, the prior distribution $\pi_{\bstheta}(\bstheta;\{\boldp_{i}\}_{i=1}^N)$ should assign a larger weight to the parameter $\bstheta$ that results in the probability function $\pi_{\boldq}(\boldq_i; \boldp_{i})$ having a high probability around $\boldp_{i}$. 
It is shown by taking $\pi_{\boldq}(\boldq;\boldp_{i})$ as the density function of a Dirichlet distribution $Dir(\mathbf{1}_K+\lambda \boldp_{i})$, where $\lambda$ is a tuning parameter controlling the confidence in the predicted probabilities of the teacher model $M_t$, the optimization of maximizing the posterior distribution of $\bstheta$ is equivalent to solving the objective function of KD.

Another advantage of BKD lies in its straightforward uncertainty quantification, which is achieved via the posterior distribution of the student model's weights.
Within this framework, a suite of Bayesian inference tools enables posterior sampling for uncertainty quantification in the student model. For example, the stochastic Gradient Langevin Dynamics (SGLD) \citep{welling2011bayesian} can efficiently sample the student model's weights from the derived posterior distribution. 
Using these posterior samples, we can compute the expected value of predictive evaluation metrics, including variance in regression or deviance in classification, by computing their posterior mean.
This could provide a robust estimate of uncertainty in the population.

\paragraph{Challenges and Future Directions.} Uncertainty quantification typically incurs significantly higher computational costs than standard prediction tasks, particularly when relying on large-scale sampling or Monte Carlo methods. In contrast, variational inference has demonstrated strong performance as an efficient alternative to sampling-based methods in various settings \citep{blei2017variational}. A promising yet underexplored direction is the systematic integration of variational inference into knowledge distillation frameworks to balance accuracy and computational efficiency. Meanwhile, the Bayesian framework for knowledge distillation opens up rich possibilities for advancing uncertainty quantification in this domain. Key future directions include (1) designing more flexible teacher-informed priors that incorporate statistical justification and (2) developing scenario-specific adaptations that account for different data distributions or task requirements. 

These research directions would help solidify the theoretical foundations of uncertainty-aware knowledge distillation while enhancing its practical applicability.

\subsection{Multi-Teacher KD}\label{HaoranLu}

Multi-teacher knowledge distillation offers a powerful extension to the single-teacher paradigm by consolidating heterogeneous expertise from multiple teacher models into a single student. Conceptually, it aims to utilize the diversity of teacher networks to yield richer supervision and improved generalization. The key challenge is to fuse potentially conflicting signals from distinct teachers in a principled manner.
\begin{figure}[h]
    \centering
    \includegraphics[width=0.8\linewidth]{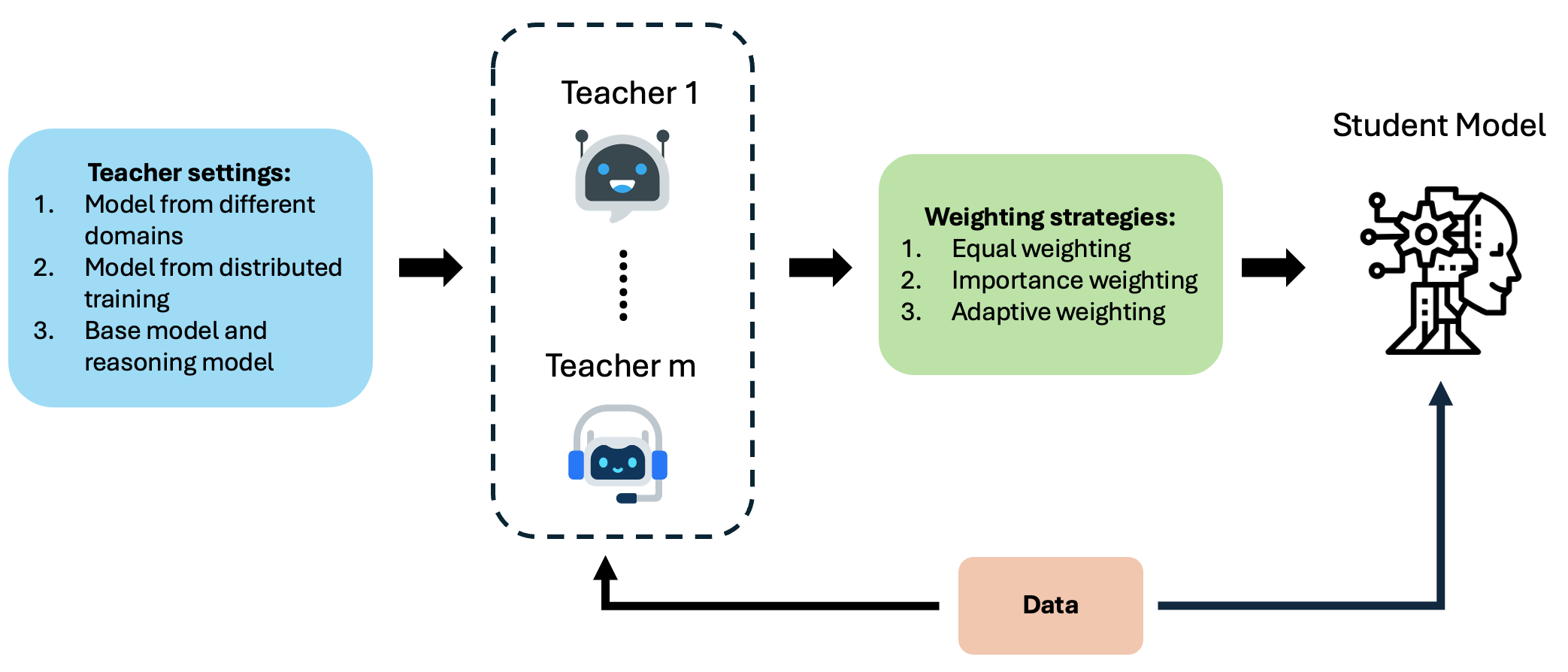}
    \caption{Multi-Teacher Knowledge Distillation. The variants of Multi-Teacher KD include different weighting strategies and teacher settings.}
    \label{fig:mt}
\end{figure}
Formally, suppose we have $K$ trained teacher models $T_{k}$ (where $k \in \{1,\ldots,K\}$). For a given input $\mathbf{x}$, each teacher produces a predictive distribution $\boldsymbol{p}_k(\boldsymbol{x})$. A simple ensemble strategy averages over these distributions:
\begin{equation}
\label{eq:multi-teacher-distill-average}
\hat{\boldsymbol{p}}(\boldsymbol{x}) \;=\; \frac{1}{K}\sum_{k=1}^{K} \boldsymbol{p}_k(\boldsymbol{x})\,.
\end{equation}

Then, the student model $S$ with parameters $\theta$ can be updated by minimizing a distillation loss, for example KL divergence, between $\hat{\boldsymbol{p}}(\boldsymbol{x})$ and the student’s output $\boldsymbol{p}_S(\boldsymbol{x}; \theta)$:
\begin{equation}
\label{eq:kl-loss}
\mathcal{L} \;=\; D_{\text{KL}}\!\Big(\hat{\boldsymbol{p}}(\boldsymbol{x}) \;\|\; \boldsymbol{p}_S(\boldsymbol{x};\theta)\Big),
\end{equation}
where $D_{\text{KL}}(a \parallel b)$ denotes the KL divergence of $a$ and $b$.
While this uniform average provides an effective first step, more sophisticated methods address the variability in teacher quality and teacher conflicts. Some refined approaches take different weighting strategies for the teachers. For example, \citet{zhang2022confidence} assigns large weights to teacher predictions that closely match the ground truth, thus mitigating the adverse influence of unhelpful teachers. On the other hand, adaptive weighting has also been investigated to ensure a more dynamic synthesis of teacher signals. \citet{du2020agree} explored an adaptive strategy that merges multi-teacher outputs in gradient space, making it possible to resolve conflicts without discarding the diversity they offer. 

Further, instead of using a simple loss, such as the mean squared error, \citet{you2017learning} proposed aligning the student's feature embeddings with those of the teachers via a distance metric, capturing complementary knowledge.

Recent research has focused on using different teacher settings in multi-teacher KD to enhance specific skills.
\citet{tian2024beyond} proposed TinyLLM, a multi-teacher KD method that transfers reasoning skills from large LLMs to smaller ones by not only focusing on correct answers but also capturing the underlying rationales. TinyLLM uses an in-context example generator and a teacher-forcing CoT strategy to ensure reasoning accuracy and diversity of knowledge sources. The method demonstrated significant performance improvements across reasoning tasks by effectively synthesizing the diverse problem-solving strategies of different teachers.
\citet{khanuja2021mergedistill} introduced MERGEDISTILL, which merges pre-trained multilingual and monolingual LLMs into a single task-agnostic student model. The framework employs offline teacher evaluation and vocabulary mapping to integrate knowledge from diverse models, enabling the student model to benefit from both multilinguality and the specialized knowledge of individual languages. MERGEDISTILL's approach is particularly effective in scenarios with overlapping language sets, showing that leveraging diverse teacher models can create a robust and versatile student model.
\citet{liu2024wisdom} developed DIVERSEDISTILL, a framework that addresses the challenge of teacher-student heterogeneity by introducing a teaching committee comprising various teachers. The method dynamically weights teacher contributions based on the student's understanding of each teacher's expertise, which helps in scenarios with significant architectural or distributional differences among the teachers. The method is highly adaptable and has demonstrated improved performance in both vision and recommendation tasks.
Lastly, other than focusing on the training process, \citet{wadhwa2025taught} explored an interesting concept of `teacher footprints' in distilled models, proposing methods to identify which teacher model was used to train a student LLM. Their work emphasizes the importance of understanding teacher influence, particularly when using proprietary models, and developed discriminative models based on syntactic patterns to trace teacher origins.

\paragraph{Challenges and Future Directions.}
Multi-teacher KD for LLMs introduces new complexities beyond the traditional single-teacher paradigm. On one hand, integrating expertise from multiple teachers can greatly enrich a student’s generalization and domain coverage, especially when those teachers specialize in different areas (e.g., biomedical vs. legal). However, practical constraints arise in handling teacher availability: many approaches assume full access to all teacher models, which is often unfeasible due to licensing, privacy, or asynchronous deployment. Moreover, distilling from multiple teachers incurs high computational overhead, pairwise interactions among $K$ teachers scale roughly as $O(K^2)$, making it hard to synchronize their distinct objectives and representations. Aligning knowledge across teachers with heterogeneous architectures or tokenization schemes also demands careful design to reconcile discrepancies. A straightforward research step to probe these issues is to evaluate multi-teacher students on domain-specific benchmarks, ensuring that each teacher’s expertise is retained. For instance, if one teacher excels in biomedical QA and another in legal reasoning, the student could be tested on a biomedical QA dataset and a legal reasoning dataset to verify it meets or exceeds each teacher’s performance in-domain. Such experiments would reveal whether multi-teacher distillation truly broadens the student’s skill set or whether some knowledge gets diluted.

Another key challenge is managing conflicts when teachers produce divergent or contradictory outputs. Without careful resolution, the student may either homogenize the teachers’ errors or fail to benefit from each teacher’s unique perspective. While various weighting and ensemble strategies attempt to reconcile teacher signals, they remain limited in open-ended LLM scenarios where disagreements can be subtle or domain-specific. An important open question is how a student can detect and resolve contradictory guidance from different teachers. One possible experimental setup to explore this would be to inject intentionally conflicting answers from teachers for certain training queries and then evaluate different conflict-resolution strategies (e.g., confidence-based weighting, context-dependent teacher selection) on held-out test cases to see which yields the most accurate and consistent student responses. In addition, tracing or quantifying each teacher’s individual contribution to the student is non-trivial, especially in collaborative, multi-domain settings where teachers’ knowledge areas overlap or their stylistic preferences blur together. Recent work by \citet{wadhwa2025taught} takes an initial step toward this by identifying teacher footprints in the student, but we lack general methods to robustly attribute portions of the student’s knowledge to specific teachers. A concrete research question here is whether we can develop interpretability tools or influence-function analyses to trace knowledge transfer: for example, performing leave-one-teacher-out ablation studies (training students while withholding one teacher at a time) and measuring performance drops in that teacher’s specialty could quantify its influence. Developing such diagnostics would not only strengthen trust in distilled models but also guide the design of better fusion mechanisms. Ultimately, creating lightweight yet robust methods to fuse multi-teacher signals, resolve domain conflicts, and accommodate partial teacher availability stands as a central objective for advancing multi-teacher KD in LLMs. By formulating explicit evaluation protocols, such as cross-domain performance checks and contradiction-handling tests, future research can make multi-teacher distillation more principled and actionable.

\subsection{Dynamic and Adaptive KD}

Previous approaches rely on a static, pre-trained teacher LLM to transfer knowledge to a student LLM. Dynamic and adaptive approaches introduce a paradigm shift by fostering a bidirectional collaboration between models. These frameworks emphasize two key strategies: (1) simultaneous co-evolutionary training, where teacher and student are both refined during joint optimization, and (2) self-distillation, which bypasses the need for a pre-trained teacher entirely by enabling a single model to generate and distill its own knowledge. By departing from fixed hierarchies, such methods enhance adaptability, mitigate the limitations of static teacher expertise, and improve generalization in evolving or resource-constrained scenarios.

\subsubsection{Simultaneous Training of Teacher and Student model}\label{Xilin}

Simultaneous training of teacher and student models in knowledge distillation is an approach where both models are optimized together during the training process, rather than using a pre-trained teacher model to guide the student passively \citep{sun2021collaborative, chang-etal-2022-one}. This approach allows for dynamic interaction, where the student continuously refines its learning based on real-time feedback from the teacher, and the teacher can adapt its guidance based on the student's progress.

Some methods incorporate constraints in the loss function to facilitate joint learning of the student and teacher models. For instance, \citet{li2024bild} proposed a new loss function named Bi-directional Logits Difference. Instead of directly aligning their logits, it computes pairwise differences between the top-k logits for both models. This results in two types of differences: the teacher-led logits difference (t-LD) loss, which captures key knowledge from the teacher’s top-k logits, and the student-led logits difference (s-LD) loss, which forces the teacher to consider the student’s perspective. The BiLD loss is given by 
\begin{equation}
    L_{\text{BiLD}} = D_{\text{KL}}(\boldsymbol{p}_{t-LD} \parallel \boldsymbol{p}_{s-LD}) + D_{\text{KL}}(\boldsymbol{p}_{s-LD} \parallel \boldsymbol{p}_{t-LD}) ,
\end{equation}
where $D_{\text{KL}}(a \parallel b)$ denotes the KL divergence of $a$ and $b$, $\boldsymbol{p}_a$ denotes the probability distribution of $a$.
 
\begin{figure}[h]
    \centering
    \includegraphics[width=0.8\linewidth]{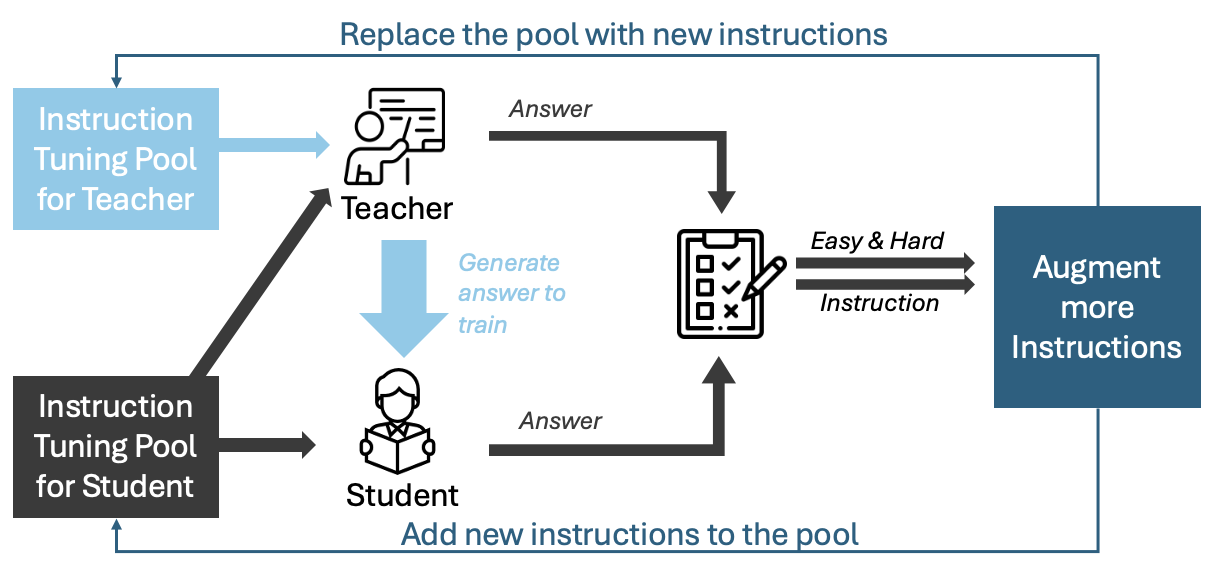}
    \caption{Competitive Multi-modal Distillation Framework for Joint Training of Teacher and Student Models. It includes instruction tuning, evaluation, and augmentation phases, with a bidirectional feedback loop to iteratively improve both models. The assessor model guides instruction refinement based on performance comparison.}
    \label{fig:joint_training}
\end{figure}

Other methods employ specially designed training frameworks to enable joint learning. \citet{li2023unlock} introduced a Competitive Multi-modal Distillation framework that introduces a bidirectional feedback loop through multi-modal competitive distillation, the overall framework is shown in Figure \ref{fig:joint_training}. It consists of three key phases: instruction tuning, instruction evaluation, instruction augmentation. In the first phase, the teacher model is first tuned using its designated instructions before generating responses for questions from the instruction tuning pool. The student model is then trained to align its responses with the teacher’s outputs.  In the evaluation phase, both the teacher and student models generate responses for the instructions, which are then evaluated by an assessor model. Finally, in the augmentation phase,  new instructions are generated, then replace or add to the original instruction pool.

\subsubsection{Self-distillation} \label{yufang}

\begin{figure}[h]
    \centering
    \includegraphics[width=0.95\linewidth]{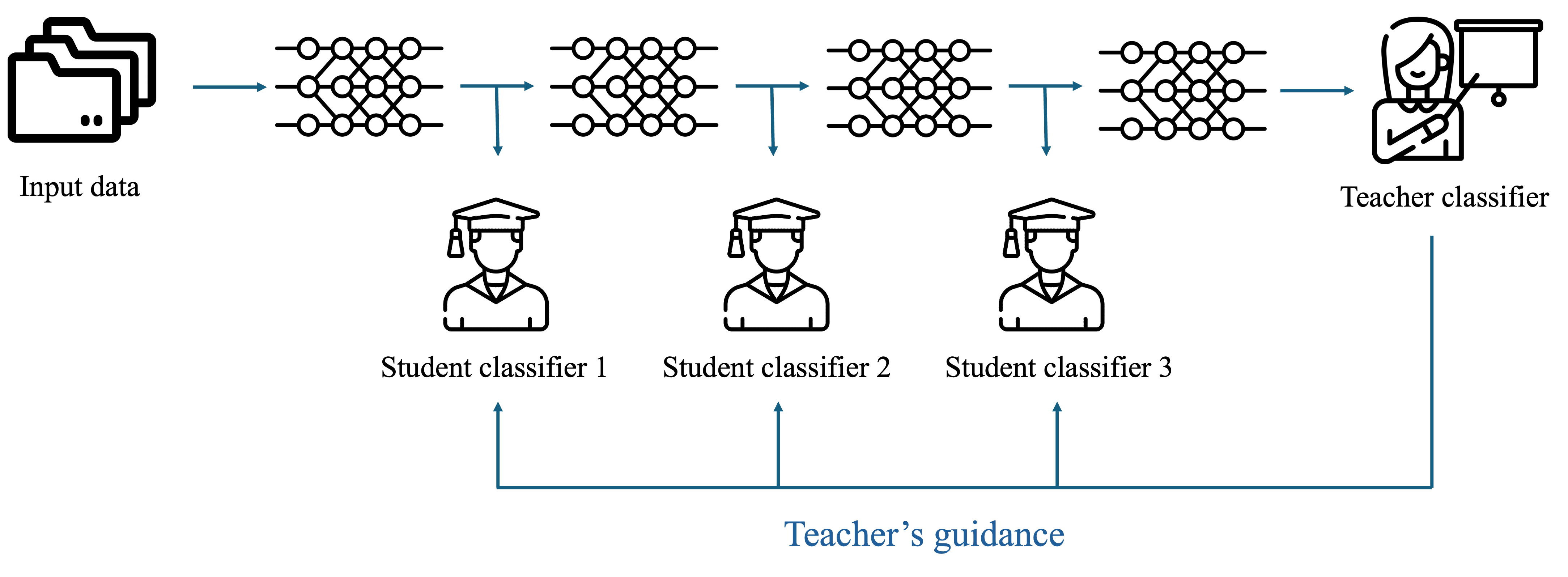}
    \caption{Framework for Self-Distillation. The model is partitioned into sequential sections, each followed by a shallow student classifier. The deepest classifier acts as the teacher, guiding earlier classifiers through knowledge distillation.}
    \label{selfkd}
\end{figure}

In self-distillation, instead of using a separate teacher model to guide students, the model is trained so that the teacher and students learn simultaneously. The overall framework is shown in Figure \ref{selfkd}. The architecture is divided into multiple blocks, with intermediate outputs extracted from earlier blocks. These outputs pass through an attention module and are fed into a shallow classifier, which makes predictions based on only the initial blocks. In contrast, the final classifier utilizes all blocks before making predictions. Here, the final classifier acts as the ``teacher'', while the shallow classifiers serve as the ``students'' \citep{Zhang2019BeYO}. 

The teacher's guidance in self-distillation is enforced through a carefully designed loss function composed of three components. The first is the cross-entropy loss between the true labels and each classifier’s predictions. The second measures the KL divergence between the softmax output probabilities of each shallow (student) classifier and the final (teacher) classifier. The third imposes an L2-norm penalty on the difference between the features fed into each shallow classifier and those used by the final classifier.

In addition to this three-part loss function, various loss function designs capture the relationships between students and teachers. For instance, rather than computing KL divergence solely between the final classifier and each shallow classifier, it can be computed iteratively: first between the final classifier and its predecessor, then progressively between successive classifiers moving backward through the network. This approach is known as transitive teacher distillation. Alternatively, instead of measuring the KL divergence between each shallow classifier and the final classifier, it can be computed between each classifier’s output and the ensemble prediction of all classifiers, a technique called ensemble teacher distillation \citep{Zhang2022SelfDistillation}. Furthermore, \citet{DBLP:journals/corr/abs-2012-09816} provides a theoretical guarantee that self-distillation with an ensemble of independently trained neural networks can improve test accuracy when data exhibits a multi-view structure, reinforcing the understanding that the dark knowledge embedded in ensemble outputs contributes to model generalization.

In the development of AlphaFold, self-knowledge distillation played a pivotal role in enhancing the model's accuracy. Initially, AlphaFold was trained using known protein structures from the Protein Data Bank. Subsequently, the model predicted structures for approximately 300,000 diverse protein sequences obtained from Uniclust. These predictions were then incorporated back into the training process, allowing the model to learn from its own predictions. This approach effectively utilized unlabeled sequence data, leading to significant improvements in the model's performance \citep{jumper2021highly}.

Self-distillation technique is particularly valuable as it enables the model to leverage vast amounts of unlabeled data, overcoming the limitations imposed by the availability of labeled training data. By refining its predictions through iterative learning from its own outputs, AlphaFold achieved unprecedented accuracy in protein structure prediction \citep{yang2023alphafold2}.

\subsubsection{Challenges and Future Directions}
While dynamic and adaptive KD enables flexible collaboration between models, practical challenges remain. First, simultaneous training of teacher and student models risks unstable convergence due to interdependent learning dynamics, requiring carefully designed loss functions (e.g., bidirectional feedback) to prevent conflicting objectives. 
Second, self-distillation avoids reliance on external teachers but can amplify inherent model biases or errors when its guidance lacks calibration. It is necessary to develop automatic bias detectors that flag flawed knowledge and enable self-correction through adversarial questioning or consistency checks across multiple inference paths. Another promising direction is to add minimal external supervision, such as human-annotated bias signals and uncertainty-aware loss functions, to recalibrate guidance without sacrificing efficiency.
Third, real-time interactions between models significantly increase computational overhead, posing greater demands on computational resources. Future research may focus on adaptive training protocols that dynamically balance knowledge exchange, leveraging lightweight feedback or noise-robust distillation strategies. 
Iterative self-correction using both labeled and unlabeled data, as exemplified by AlphaFold, may further improve reliability. Enhancing computational efficiency through modular architectures or sparse interaction mechanisms could increase scalability. Finally, developing interpretable metrics to quantify knowledge transfer would help build trust in dynamic KD systems.

Some existing work has proposed potential solutions. SCKD \citep{SCKD} and ISD-QA \citep{isdqa} use iterative self-correction with unlabeled data to enhance robustness. SparseBERT \citep{SPARSEBERT} improve scalability through modular or sparse designs. Proto2Proto \citep{PRO2PRO} and SSCBM \citep{husscbm} develop interpretable metrics to quantify and visualize knowledge transfer. While these methods offer valuable insights and techniques, they have yet to be extensively explored or applied to LLMs, highlighting an important direction for future research.


\subsection{Vision KD for Autonomous Driving}
\label{subsec:vlm_ad_distillation}

Recent advancements in Large Vision-Language Models (VLMs) have demonstrated exceptional capabilities in semantic scene understanding, causal reasoning, and open-vocabulary object detection. In the domain of Autonomous Driving (AD), these capabilities are crucial for handling complex, long-tail scenarios that traditional geometric planners fail to address (e.g., interpreting hand gestures, navigating construction zones, or reasoning about occluded agents). However, the high computational latency and memory footprint of billion-parameter VLMs prohibit their direct deployment on resource-constrained vehicle edge devices. Consequently, a new paradigm of \textit{Vision-Language Distillation} has emerged to transfer the reasoning capabilities of VLMs into lightweight, real-time driving policies.

Unlike traditional KD which typically focuses on logit matching, vision distillation in AD requires transferring spatial-temporal awareness and causal logic. We categorize these methodologies into three primary strategies: Rationale-Based Feature Alignment, Surrogate Task Distillation, and Contrastive Planning.

\begin{table*}[h]
\centering
\caption{Comparison of distillation methodologies transferring knowledge from VLMs to autonomous driving agents.}
\label{tab:vlm_ad_distillation}
\resizebox{\textwidth}{!}{
\begin{tabular}{@{}lllll@{}}
\toprule
\textbf{Method} & \textbf{Teacher Model} & \textbf{Student Model} & \textbf{Knowledge Type} & \textbf{Distillation Mechanism} \\ \midrule
\textbf{VERDI} \citep{feng2025verdi} & VLM (e.g., Qwen-VL) & End-to-End Planner & Causal Rationales & Latent Space Feature Alignment \\
\textbf{DiMA} \citep{hegde2025distilling} & Multi-modal LLM & Vision Planner & World Knowledge & Joint Training w/ Surrogate Tasks \\
\textbf{VLP} \citep{pan2024vlp} & Large Language Model & BEV Backbone & Semantic Prototypes & Agent-Centric Contrastive Learning \\
\textbf{Vi-LAD} \citep{elnoor2025vilad} & VLM & Navigation Policy & Social Attention & SSIM Loss on Attention Maps \\
\textbf{DSDrive} \citep{liu2025dsdrive} & VLM (CoT) & Compact LLM & Reasoning + Waypoints & Dual-Head Coordination Loss \\
\textbf{Lingo-1} \citep{wayve2023lingo} & Expert Driver + LLM & VLA Model & Explanations & Action-Commentary Generation \\ \bottomrule
\end{tabular}
}
\end{table*}

\subsubsection{Rationale-Based Feature Alignment}
Driving decisions are often driven by causal logic (e.g., ``I am stopping \textit{because} the pedestrian is distracted"). \citet{feng2025verdi} introduces a framework, \textbf{VERDI} (VLM-Embedded Reasoning for Autonomous Driving), where a VLM teacher generates textual rationales for ground-truth trajectories. These rationales are encoded into a semantic latent space. The student model is trained to align its intermediate feature maps (from perception, prediction, and planning modules) with these linguistic embeddings via a learnable projector $\mathcal{P}$. The alignment loss is formulated as:
\begin{equation}
    \mathcal{L}_{align} = \sum_{k \in \{per, pred, plan\}} \left( 1 - \cos(\mathcal{P}_k(F_k), E_{text}(R)) \right)
\end{equation}
where $F_k$ represents the student's feature map at stage $k$, and $E_{text}(R)$ is the embedding of the VLM-generated rationale $R$. This forces the student to organize its latent space semantically, improving robustness in zero-shot scenarios. Similarly, \textbf{DSDrive} \citep{liu2025dsdrive} employs a waypoint-driven dual-head coordination module to synchronize the distilled reasoning with kinematic planning outputs.

\subsubsection{Surrogate Task Distillation}
To enforce deeper scene understanding, the \textbf{DiMA} (Distilling Multi-modal LLMs) framework \citep{hegde2025distilling} utilizes a joint training strategy where a shared Scene Encoder serves as a tokenizer for an MLLM teacher. The framework introduces surrogate tasks such as \textit{Masked Token Reconstruction} and \textit{Future Prediction}. The gradients from the MLLM's reasoning objectives backpropagate into the shared encoder, enriching the representations used by the lightweight planner. The reconstruction objective is defined as:
\begin{equation}
    \mathcal{L}_{recon} = |
| \hat{B}_{masked} - B_{gt} ||_2^2
\end{equation}
where $\hat{B}_{masked}$ represents the reconstructed Bird's-Eye-View (BEV) tokens. This enables the student planner to leverage the MLLM's world knowledge during training while remaining independent at inference time.

\subsubsection{Contrastive Vision-Language-Planning (VLP)}
The \textbf{VLP} framework \citep{pan2024vlp} addresses generalization by employing contrastive learning to align visual scene elements with linguistic prototypes. It introduces an \textit{Agent-centric Learning Paradigm (ALP)}, where cropped BEV features of scene agents (vehicles, pedestrians) are aligned with their text descriptions via an InfoNCE loss:
\begin{equation}
    \mathcal{L}_{VLP} = - \mathbb{E} \left[ \log \frac{\exp(sim(f_{bev}, f_{text})/\tau)}{\sum_{neg} \exp(sim(f_{bev}, f_{neg})/\tau)} \right]
\end{equation}
This ensures that the perception backbone learns semantically distinct representations for long-tail objects, significantly reducing collision rates in unseen environments. Other approaches, such as \textbf{Vi-LAD} \citep{elnoor2025vilad}, extend this by distilling attention maps to ensure socially compliant navigation behaviors.

\vspace{0.3cm}
\noindent
\textbf{Summary.} The autonomous driving domain exemplifies why Knowledge Distillation has become indispensable in modern vision systems. Unlike NLP tasks where inference can tolerate higher latency, vision applications, particularly safety-critical scenarios like AD, demand real-time decision-making under strict hardware constraints (e.g., $<$50ms latency on vehicle edge devices). VLMs, while achieving remarkable semantic reasoning capabilities, are fundamentally incompatible with these deployment requirements due to their billion-parameter architectures. KD bridges this gap by distilling the reasoning depth, spatial-temporal awareness, and causal logic of VLMs into compact student models that preserve critical capabilities while achieving orders-of-magnitude speedup. The methodologies reviewed here, including rationale-based alignment, surrogate task learning, and contrastive planning, demonstrate that effective distillation in vision extends beyond simple logit matching and requires transfer of structured spatial knowledge and semantic prototypes. This paradigm has broad implications beyond Autonomous Driving: medical imaging diagnosis, robotic manipulation, and video understanding all face similar trade-offs between model capacity and deployment feasibility, positioning vision-centric KD as a foundational technique for practical AI systems.

\subsection{Task-Specific KD}\label{Xilin}

Task specific distillation is often combined with instruction tuning, where a large instruction-tuned teacher model distills its task-specific knowledge into a smaller student model while preserving instruction-following capabilities \citep{wu2023lamini, yang2023enabling, zhang2023instruction}.

\citet{taori2023alpaca} and \citet{wei2021finetuned} introduced instruction-tuned LLMs that distill task-specific knowledge through instruction-following datasets. Alpaca fine-tunes Meta's LLaMA-7B using 52,000 instruction-response pairs generated by OpenAI’s text-davinci-003, mimicking knowledge distillation. Similarly, FLAN fine-tunes a 137B model on over 60 NLP datasets phrased as natural language instructions, enabling better generalization to unseen tasks. Both methods leverage instruction tuning to enhance model performance across diverse domains, outperforming larger models like GPT-3, LaMDA-PT, and GLaM in tasks such as inference, question answering, and translation.

Although LLMs can achieve high performance through instruction tuning, the predefined instructions are often simpler than the complex cases encountered in real-world scenarios \citep{zhang2023alpacare, wang2022self,ouyang2022training}. Therefore, domain-specific distillation, which tailors knowledge transfer for specialized fields (e.g.medical, programming), is crucial for improving model effectiveness. 
Most solutions to these problems involve designing a domain-specific dataset to fine-tune the model. \citet{zhang2023alpacare} proposed AplaCare,  which enhances medical LLMs through a semi-automated instruction fine-tuning pipeline. Starting with a clinician-curated seed dataset, they used GPT-4 to generate diverse medical tasks and ChatGPT to create responses, forming MedInstruct-52k. This dataset fine-tunes LLaMA models, improving instruction-following ability. Their approach achieves significant gains in medical benchmarks. 

\subsection{Theoretical Studies}\label{yongkai}
Theoretical studies of knowledge distillation focus on explaining how knowledge is transferred and why this process is effective.
The understanding is built upon several main insights, including (1) the interpretation of the teacher-student paradigm, (2) the convergence and generalization theory, (3) the impact of architectural choices on knowledge transfer, and (4) the role of data in knowledge distillation.

The key aspect of the teacher-student paradigm is the use of soft labels, as introduced in Section \ref{sec:Def}.
One interpretation is that these soft labels contain dark knowledge, which reveals information about both the teacher's uncertainty and the relationships between classes \citep{muller2019does}. 
In addition, several studies consider the soft label as label smoothing regularization, which provides the trade-off between bias and variation when training the student model \citep{yuan2020revisiting,zhou2021rethinking}. More recent studies provide the systematic Bayesian framework for interpreting the regularization of the knowledge distillation loss as the effect of implicitly imposing a prior distribution on the student model \citep{menon2021statistical,fangbayesian2024}. Furthermore, the temperature paper $\tau$ in Equation (\ref{eq: KD loss}) can be interpreted as the variance parameter of the prior or as a measure of dispersion within the framework of statistical Bayesian modeling 
 \citep{fangbayesian2024}.

Several theoretical studies have examined the convergence behavior of student models in the context of knowledge distillation.
Most research often focuses on simplified model architectures such as linear models, deep linear models, or the Gaussian process
\citep{phuong2019towards,mobahi2020self,borup2021even}.
For more general cases, \citet{menon2021statistical} establishes a concrete criterion for evaluating the performance of a teacher model. It is proven that knowledge distillation can reduce the prediction variance of the student model, ultimately resulting in a more accurate student model.
Some studies have surprisingly indicated that a student model trained via distillation can obtain good generalization bounds even from a teacher model that having poor theoretical bounds \citep{hsu2021generalization}.
This is related to the regularization effect of KD. The model compression of knowledge distillation can also be viewed as a form of regularization of the model space, which leads to better performance on the testing
data.
Furthermore, \citet{lopez2015unifying} unifies the knowledge distillation with privileged information \citep{vapnik2015learning} and proposes the generalized distillation. The benefits of learning with a teacher are shown to arise due to three factors: (1) The capacity of the teacher model is sufficiently small, allowing it to be well approximated by a smaller student model.
(2) The teacher's approximation error relative to the true or optimal decision boundary is smaller than the student's approximation error.
(3) The rate at which the student learns the teacher's decision boundary is faster than the rate at which the student learns the true function. However, providing a rigorous justification for these conditions remains an open question in the field.

The difference in model capacity between the teacher and the student models, often referred to as the capacity gap, is a crucial factor influencing the effectiveness of distillation \citep{mirzadeh2020improved,li2023kd,niu2022respecting}. 
A large capacity gap, where the student might lack the representational power to fully absorb the knowledge learned by the teacher, can sometimes hinder effective knowledge transfer. Conversely, if the capacity gap is too small, the student might already have sufficient capacity to learn the task effectively from the original data, making distillation less beneficial.
\citet{zhang2023towards} finds that the optimal teacher scale almost consistently follows a linear scaling with the student scale across different language model architectures and data scales in terms of the scaling law.

The choice of the distillation dataset can also influence what aspects of the teacher's knowledge are learned by the student. Increasing the size of the distillation dataset does not always lead to improved student fidelity to the teacher and can even negatively affect the student's generalization performance \citep{stanton2021does}.
\citet{phuong2019towards} studied the effect of data geometry on knowledge distillation in the case of linear and deep linear models. They describe the data geometry by defining the angular alignment between the angular alignment between the data teacher's weight vector, which has been proven to be a crucial factor in the derived generalization
bound.

\begin{table*}[htbp]
\centering
\caption{Comparative Summary of Rrepresentative KD Methods for LLMs.}\label{tab:compare_KD}
\resizebox{0.95\textwidth}{!}{
\begin{tabular}{p{4.5cm}p{3.5cm}p{3.5cm}p{3.5cm}}
\toprule
\textbf{KD paradigm} & \textbf{Key idea / strengths} & \textbf{Typical limitations} & \textbf{Representative use cases} \\
\midrule
Rationale-based KD \citep{hsieh2023distilling,chu2023survey,feng2024keypoint} & Transfers chain-of-thought rationales, yielding interpretable students with strong generalization and lower data demand & Needs reliable rationales and metrics; harder to extract rationale knowledge with black-box teachers & Decision-critical domains (clinical or legal reasoning) \\
\midrule
Uncertainty-aware KD \citep{korattikara2015bayesian,vadera2020generalized,malinin2019ensemble,menon2021statistical,fangbayesian2024} & Distils predictive distributions, giving calibrated confidence and robustness to noise & Bayesian sampling / ensemble distillation adds computational cost & Safety-critical or noisy-label tasks (medical diagnosis, risk assessment) \\
\midrule
Multi-teacher KD \citep{you2017learning,zhang2022confidence,du2020agree,fukuda2017efficient,zhu2021data,tian2024beyond,khanuja2021mergedistill,liu2024wisdom,wadhwa2025taught} & Fuses complementary expertise from several teachers to broaden coverage and robustness & High compute cost; reconciling conflicts among heterogeneous teachers & Cross-domain or multilingual students inheriting specialized skills \\
\midrule
Dynamic and adaptive KD \citep{chang-etal-2022-one,sun2021collaborative,li2024bild,li2023unlock,Zhang2019BeYO,Zhang2022SelfDistillation,DBLP:journals/corr/abs-2012-09816,jumper2021highly,yang2023alphafold2} & Teacher and student co-evolve, or a model refines itself, enabling continual learning with vast unlabeled data & Possible unstable convergence, bias amplification, and training cost & Rapidly evolving fields or unlabeled-data regimes (e.g., AlphaFold protein prediction) \\
\midrule
Task-specific KD \citep{wu2023lamini,yang2023enabling,zhang2023instruction,taori2023alpaca,wei2021finetuned,zhang2023alpacare, wang2022self,ouyang2022training} & Tailors transfer via instruction tuning, giving compact students strong task performance with minimal data & Requires curated instruction data; simple instructions may miss real-world complexity & Domain specific applications, such as medical LLMs, specialist code or legal assistants \\
\bottomrule
\end{tabular}
}
\end{table*}

\subsection{Summary and comparison of KD methods}\label{sec:KD_summary}

In prior subsections, we examined the principal KD methodologies in detail. Table \ref{tab:compare_KD} then offers a concise overview of each method’s core idea, its main strengths and limitations, and representative use cases. This summary serves as a quick reference to guide readers in choosing the most appropriate approach based on model capacity, computational budget, and application domain.

To give a complementary perspective, Table \ref{tab:KD_compare_vs} contrasts single-teacher and multi-teacher KD, and domain-specific and general-purpose KD. It outlines each method’s key idea, data requirements, robustness/performance, computational cost/flexibility, and typical application scenarios, helping practitioners identify the approach best suited to their objectives.

\begin{table*}[ht]
\centering
\caption{Comparative overview of single-teacher versus multi-teacher KD and domain-specific versus general-purpose KD.}\label{tab:KD_compare_vs}
\footnotesize
\begin{tabular}{p{2.2cm}p{4.5cm}p{4.5cm}}
\toprule
\textbf{Aspect} & \textbf{Method A} & \textbf{Method B} \\
\midrule
\multicolumn{3}{l}{\textit{Single-teacher vs Multi-teacher KD}} \\
\midrule
Key idea & Distill knowledge from one expert teacher & Fuse knowledge from an ensemble of multiple teachers \\
Data requirement & Outputs from a single teacher model & Outputs from several teacher models, increasing diversity \\
Robustness & Performance tied to one teacher’s biases & Improved generalization by reconciling complementary expertise \\
Compute cost & Moderate (one forward pass per example) & Higher (multiple forward passes per example) \\
Use cases & When a single strong teacher is available or compute is constrained & Cross-domain transfer, multilingual KD, ensemble compression \\
\midrule
\multicolumn{3}{l}{\textit{Domain-specific vs General-purpose KD}} \\
\midrule
Key idea & Tailor distillation on specialized, domain-relevant data & Distill on broad, diverse corpora for wide applicability \\
Data requirement & High-quality, annotated domain data & Large-scale, heterogeneous datasets covering many topics \\
Performance & Superior in the target domain & Good across varied tasks but may underperform on niche tasks \\
Flexibility & Limited outside the trained domain & High adaptability to new tasks with minimal re-distillation \\
Use cases & Clinical decision support, legal analysis, scientific literature & Open-domain chatbots, general question answering, conversational agents \\
\bottomrule
\end{tabular}
\end{table*}

\section{Methodologies and Techniques for Dataset Distillation in LLMs}\label{sec:method_DD}

LLMs are trained on massive corpora that often include redundant, low-quality, or uninformative samples. To improve training efficiency without sacrificing performance, dataset distillation techniques aim to synthesize compact datasets that retain the essential information of the full corpus. These methods generally fall into two main categories: optimization-based approaches, which directly learn a small set of synthetic samples by optimizing them to replicate the behavior of the full dataset during training; and synthetic data generation, which leverages generative models or heuristics to produce representative training examples. In addition to distillation, we also review data selection strategies, which focus on identifying high-quality subsets from existing datasets to achieve optimal model performance with fewer training samples.

\begin{figure}[!ht]
\centering
\vspace{-10pt}
\includegraphics[width=0.8\textwidth]{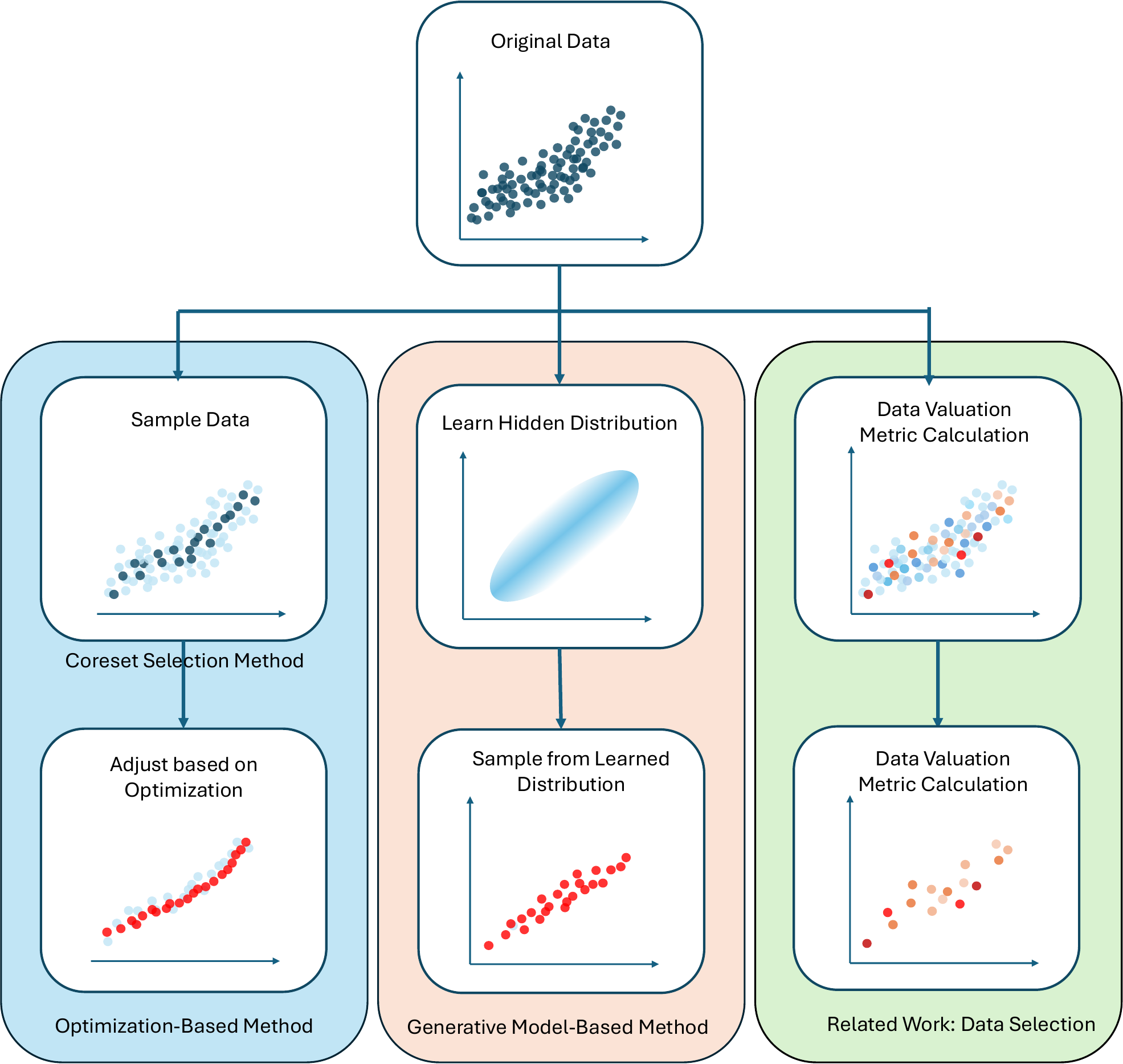}
\caption{Taxonomy of Dataset Distillation, including Data Selection, Optimization-based methods, and Generative Model-based Methods. The root plot shows the original dataset. The first column illustrates the data selection and pruning methods. Data points in different colors represent data with different importance scores. The warmer the color, the higher the importance score of the data point. After selection, data points with higher importance scores are preserved. The second column shows the optimization-based methods, which start with a subset of the original data and iteratively adjust the selected data according to an optimization loss function. The third column shows the generative model-based methods. These methods first train a generative model that learns the hidden distribution of the original data and then generates new data from the learned distribution.}\label{fig:DataDiltill}
\end{figure}

\begin{table}[h]
\centering
\caption{Advantages and Disadvantages of Different Dataset Distillation Methods.}\label{tab:data_distillation_methods}
\small 
\setlength{\tabcolsep}{4pt} 
\begin{tabular}{@{}p{2.7cm}p{3.4cm}p{3.5cm}p{2.2cm}@{}}
\toprule
\textbf{Method} & \textbf{Advantages} & \textbf{Disadvantages} & \textbf{Use Cases} \\
\midrule
\textbf{Optimization-Based Methods} (\citealp{yu2023dataset, wu2025data, liu2024noisy, zhong2024efficient}) & 
\textit{Adaptive Learning}: Continuously refines data selection. \newline
\textit{Flexibility}: Can be tailored to specific tasks. & 
\textit{Computational Demand}: Resource-intensive process. \newline
\textit{Complexity}: Requires careful hyper-parameter tuning. &
Best when GPU budget allows iterative meta-optimization. \\
\midrule
\textbf{Generative Model-Based Methods} (\citealp{sajedi2024data, li2024generative, kanagavelu2024medsynth, zhang2025survey}) & 
\textit{Data Augmentation}: Creates diverse data samples. \newline
\textit{Privacy Preservation}: Synthetic data mitigates privacy concerns. & 
\textit{Model Complexity}: Training accurate models is challenging. \newline
\textit{Quality Assurance}: Ensuring accurate representation is difficult. &
When diversity or privacy is paramount with sufficient compute. \\
\midrule
\textbf{Data Selection} (\citealp{yu2023dataset, yang2022dataset, tan2025data, marion2023less}) & 
\textit{Efficiency}: Reduces dataset size and training time. \newline
\textit{Simplicity}: Straightforward implementation. & 
\textit{Information Loss}: Risk of discarding valuable data. \newline
\textit{Static Evaluation}: May not adapt to evolving distributions. &
Ideal for rapid pre-filtering with limited compute. \\
\bottomrule
\end{tabular}
\end{table}

\subsection{Optimization-Based Dataset Distillation} \label{Jiazhang,Jing Zhang,soheil kolouri}

Optimization-based dataset distillation has emerged as a powerful method to compress large-scale datasets into compact, highly informative subsets for efficiently training LLMs. 
Given the exponential growth of pretraining corpora, storing and processing full datasets for model training is computationally prohibitive. 
Optimization-based approaches aim to synthesize a small yet representative dataset that can guide LLM training while preserving the generalization capability of models trained on much larger datasets.

The foundation of optimization-driven dataset distillation was established by \citet{wang2018dataset}, leveraging meta-learning to iteratively refine synthetic images. 
This foundational work led to multiple refinements focusing on improving data compression and training efficiency. 
A major advancement was the use of gradient-matching techniques, where synthetic datasets $\mathcal{D}_s$ are optimized to minimize the difference between the gradients computed on real and synthetic data $\mathcal{D}_r$:
\begin{equation}
\min_{\mathcal{D}_s} \sum_t \left\|\nabla_{\theta_t} \mathcal{L}\left(\mathcal{D}_s; \theta_t\right) - \nabla_{\theta_t} \mathcal{L}\left(\mathcal{D}_r; \theta_t\right)\right\|^2.
\end{equation}
Here, $\theta_t$ represents the model parameters at step $t$.
This ensures that training an LLM on $\mathcal{D}_s$ follows an optimization trajectory similar to training on the full dataset.
Techniques such as Dataset Condensation \citep{zhao2020dataset} and Differentiable Siamese Augmentation \citep{zhao2021dataset} employ this approach, aligning synthetic data with real dataset gradients to improve training efficiency.

Beyond gradient matching, embedding-based and trajectory-based methods further refine dataset compression for LLMs. 
Embedding-based methods optimize synthetic text sequences to preserve the statistical distribution of real embeddings in LLMs’ latent space, ensuring effective knowledge retention with fewer training samples \citep{zhao2023dataset}. 
Trajectory-based approaches, such as Matching Training Trajectories \citep{cazenavette2022dataset}, align synthetic text with full dataset training trajectories, enabling long-term consistency in optimization paths.

Efficiency improvements in LLM-specific dataset distillation have emerged in response to the immense computational demands of large-scale language model training. 
RFAD \citep{loo2022efficient} employs random feature approximation to reduce computational overhead, while FRePo \citep{zhou2022dataset} uses model-pooling techniques to mitigate overfitting in LLM training. 
DREAM \citep{liu2023dream} further enhances efficiency by replacing random text sampling with representative selection, significantly reducing distillation iterations.

\paragraph{Challenges and Future Directions.} 
Optimization-based dataset distillation for LLMs struggles to reconcile the competing demands of dataset compression and model generalization. Scaling these methods to modern LLMs introduces a critical bottleneck: preserving diverse linguistic patterns and factual knowledge in highly compressed synthetic datasets, which often prioritize common features at the expense of rare or nuanced content. Techniques like training trajectory matching face inherent limitations in maintaining alignment with full-dataset optimization paths over prolonged pretraining, as even minor discrepancies compound into significant divergence. Furthermore, evaluation frameworks remain inadequate, focusing narrowly on efficiency or task-specific accuracy while neglecting essential LLM capabilities such as factual coherence, reasoning, and cross-task adaptability.
Future research should address several concrete gaps: First, developing hierarchical trajectory matching that aligns synthetic data at multiple temporal scales (token-level, sequence-level, and epoch-level) rather than relying solely on gradient matching, with specific investigation into optimal weighting schemes for different trajectory components. Second, creating factual density metrics that quantify how well synthetic datasets preserve low-frequency but critical knowledge (e.g., historical dates, scientific constants, cultural references) compared to high-frequency linguistic patterns, potentially through knowledge graph-based validation. Third, establishing compositional reasoning benchmarks specifically designed for distilled datasets, testing whether models trained on synthetic data can perform multi-step logical inference and maintain consistency across related facts. 
Finally, integrating knowledge-aware constraints into distillation objectives could help preserve rare linguistic and factual patterns in compressed datasets, bridging the gap between efficiency and generalization.

\subsection{Synthetic Data Generation for Dataset Distillation}\label{Jiazhang}

While optimization-based methods refine existing datasets, generative-model-based dataset distillation leverages large pre-trained generative models to synthesize highly compact but expressive training corpora for LLMs.
Rather than directly optimizing dataset representations, these methods aim to learn a distribution over linguistic knowledge and generate synthetic text sequences that retain the essential structure and diversity of real data.

A key distinction of generative approaches is their use of latent space representations to encode knowledge.
Instead of optimizing individual data points, these methods sample latent variables $\mathbf{z}$ from a latent distribution $p(\boldsymbol{z})$ and map them to synthetic data using a learned generative function $G(\boldsymbol{z})$:
\begin{equation}
\boldsymbol{x}_s = G(\boldsymbol{z}), \quad \boldsymbol{z} \sim p(\boldsymbol{z}).
\end{equation}

Methods such as GLaD \citep{cazenavette2023generalizing} employ GAN-based priors to synthesize datasets that maintain high representational fidelity while being computationally efficient. It trains a generator $G$ in an adversarial setting against a discriminator $D$, where the objective is:
\begin{equation}
\min_G \max_D \mathbb{E}_{\boldsymbol{x}_r \sim p\left(\boldsymbol{x}_r\right)}\left[\log D\left(\boldsymbol{x}_r\right)\right] 
+ \mathbb{E}_{\boldsymbol{z} \sim p(\boldsymbol{z})}[\log (1-D(G(\boldsymbol{z})))].  
\end{equation}
This adversarial process ensures that the generated synthetic dataset is indistinguishable from real data. 

Besides, HaBa \citep{liu2022dataset} and KFS \citep{lee2022dataset} introduce latent-code-based synthesis, where a set of learned latent vectors and decoders reconstruct informative synthetic images. 
Unlike optimization-based methods, which refine dataset representations explicitly, generative models produce synthetic samples by learning underlying data distributions.
One key advantage of generative approaches is their ability to capture complex structural dependencies that might be difficult to encode explicitly in optimization-based methods. 

\paragraph{Challenges and Future Directions.}
Despite their strengths, generative approaches face unique challenges. Unlike optimization-based methods that explicitly control dataset refinement, generative models risk distribution drift, where synthetic text deviates from real-world linguistic structures. Additionally, generative dataset distillation for LLMs requires extensive model inference, making it computationally expensive for large-scale training. 
One promising future direction is developing efficient statistical drift detection frameworks that continuously monitor synthetic text quality during generation, with automatic generation stopping or correction when drift exceeds predefined thresholds.
Computational amortization methods are a promising cost-saving direction, such as cached intermediate representations or progressive distillation in which smaller models synthesize preliminary synthetic samples that are post-processed by larger models, possibly cutting the cost of inference drastically.
Additionally, hybrid optimization-generation pipelines provide opportunities to combine optimization-based methods with generative-based methods, where gradient-based methods identify optimal data characteristics (e.g., optimal token frequency distributions, key factual dependencies) which then guide generative models through structured prompting or fine-tuning.


\subsection{Related Works: Data Selection}\label{shushanwu}
While dataset distillation methods focus on synthesizing compact, informative training subsets, data selection addresses a closely related goal: optimizing training efficiency for LLMs by strategically identifying and selecting high-quality data \citep{albalak2024survey}.
Data selection can be used at various stages of LLM training, including data collection, data cleaning, data deduplication, selecting coreset data, and fine-tuning task-specific data. At the preprocessing stage, entropy-based filtering methods remove low-information or highly uncertain text, while perplexity-based filtering excludes content that is either too simple or too complex for the model. During the data selection phase, coreset selection techniques identify representative samples, and data attribution methods prioritize examples based on their relevance to specific tasks.
In this subsection, we review three major categories of data selection methods: data filtering, coreset selection, and data attribution.

\subsubsection{Data Filtering}
When preparing data to train a LLM, ensuring that the dataset is clean and diverse is critical \citep{albalak2024survey}. 
The first step of filtering uses simple but effective methods, including rule-based filtering \citep{penedo2023refinedweb, rae2021scaling}, keyword matching, and filtering spam data using the Fast Text model \citep{yao2020text}, which aims to strip personal information, inappropriate words, and offensive content. RefinedWeb \citep{penedo2023refinedweb} manually designed heuristic rules, like in-document repetition, URL ban list, and page length.

Following this, a secondary filter is applied to remove similar documents or sentences, which helps reduce redundancy by identifying near-duplicate content through techniques such as Jaccard similarity \citep{khan2024lshbloom}:
\begin{equation}
J(A, B)=\frac{|A \cap B|}{|A \cup B|},
\end{equation}\label{eq:jaccard}
where $A$ is the and $B$ is the set of the $n$-grams. An $n$-gram is a sequence of n consecutive tokens from the text. 
Jaccard similarity measures the similarity between two text documents, which can be viewed as a bag of $n$-grams.
Finally, advanced semantic analysis is performed using model embedding similarity filtering. In this step, embeddings for each document or sentence are generated using a pre-trained language model, and semantic similarities are calculated (commonly using cosine similarity) \citep{abbas2023semdedup}:
\begin{equation}
\operatorname{cosine} \operatorname{similarity}(\mathbf{a}, \mathbf{b})=\frac{\mathbf{a} \cdot \mathbf{b}}{\|\mathbf{a}\|\|\mathbf{b}\|},
\end{equation}
where $\mathbf{a}$ and $\mathbf{b}$ are embeddings learned by the model.
By setting a similarity threshold, data points that are semantically too similar are identified and one of them is removed. 
For document-level filtering, one path is to calculate the similarity between documents and remove one document that is above the threshold. SemDeDup \citep{abbas2023semdedup} deduplicates based on the similarity of the text data's embeddings from pre-trained language models. However, the semantically-based deduplication methods requires training LLM and is prohibitively expensive. Thus, it is more efficient to employ
LSHBloom \citep{khan2024lshbloom}, which identifies duplicated documents based on the Jaccard similarity calculated on the raw text content of each document.

For filtering high-quality data, perplexity-based filtering leverages a language models' own prediction confidence to assess the quality of text data. Perplexity is defined as the exponential of the average negative log-likelihood of the predicted words. For a sequence $w_1, w_2, ..., w_N$, it is computed as
\begin{equation}
\text { Perplexity }=\exp \left(-\frac{1}{N} \sum_{i=1}^N \log P\left(w_i \mid w_1, \ldots, w_{i-1}\right)\right).
\end{equation}
A higher perplexity indicates that the model is less confident in its predictions and indicates the data has lower quality \citep{wenzek2020ccnet}. 
FastText filter \citep{yan2024optimizing} trains a classifier or scoring model to decide which data to include. 
Entropy-based filtering methods eliminate low-information or highly uncertain text.
By calculating the cross-entropy loss from models trained on in-domain datasets and general-purpose datasets, Moore-Lewis selection
\citep{moore2010intelligent, axelrod2011domain} is proposed to identify in-domain data by selecting sentences that are much more predictable (i.e., lower cross-entropy) under the in-domain model compared to the out-of-domain model.

\subsubsection{Coreset Selection}
Coreset selection methods aim at reducing the computational burden of training machine learning models by condensing large datasets into representative subsets \citep{albalak2024survey} and have been used in linear regression \citep{li2024core, ma2015statistical}, and nonparametric regression \citep{meng2020more, zhang2018statistical, Xiaoxiaoasaa047}, deep learning \citep{mirzasoleiman2020coresets, fang2025spot, borsos2020coresets}, and network data analysis \citep{wu2023subsampling, liu2024graphsaid}.  Let 
$\mathcal{D}=\left\{\left(\boldsymbol{x}_i, y_i\right)\right\}_{i=1}^n$ be the training data and $y_i$ can be categorical or continuous for classification and regression. 
Coreset selection chooses a subset $S \subseteq \mathcal{D}$ of the full dataset $\mathcal{D}$ so that a cost function computed over $S$ closely approximates that computed $\mathcal{D}$ for all model parameters. One general formulation is:
\begin{equation}
S^*=\arg \min _{S \subseteq \mathcal{D},|S|=k} \sup _{\theta \in \Theta}\left|\frac{1}{|\mathcal{D}|} \sum_{\boldsymbol{x} \in \mathcal{D}} f( \boldsymbol{x}; \theta)-\frac{1}{|S|} \sum_{\boldsymbol{x} \in S} f(\boldsymbol{x};\theta )\right|
\end{equation}
where $f(\boldsymbol{x}; \theta )$ is the cost function evaluated at data point $\boldsymbol{x}$ with model parameters $\theta$, $\Theta$ is the set of all parameters over which the cost is evaluated, and $k$ is the size of the coreset.

In instructor tuning, \citet{zhang2024tagcos} presents a coreset selection approach that leverages gradient information from training samples. This method involves clustering data based on gradient similarities and selecting representative samples from each cluster. Similarly, Low-rank gradiEnt Similarity Search (LESS) also identifies the most influential data subsets on a target validation example using cosine similarity between the gradient features of the training samples and target examples \citep{xia2024less}. 

In the alignment step,  the coreset has to be aligned with human instructions. Their methodology focuses on evaluating data samples across three key dimensions: complexity, quality, and diversity \citep{liumakes}. They start by selecting data samples with higher scores of complexity and quality of instruction-response pairs.
To ensure diversity, they select samples that are sufficiently different based on the cosine distance of the embeddings.

In task-specific fine-tuning, STAFF \citep{zhangstaff} addresses the challenges of computational overhead and data relevance. The method leverages a smaller model from the same family as the target LLM to speculate on data importance scores efficiently. These speculative scores are then verified on the target LLM to accurately identify and allocate more selection budget to important regions while maintaining coverage of easier regions.

\subsubsection{Data Attribution}
Data attribution methods select data by quantifying the value of individual data points. The concept of data Shapley \citep{ghorbani2019data} and its variants \citep{wang2024data, wangrethinking} provide a tool for explaining the importance of each data point. The Shapley value, defined based on the prediction and the performance score of the predictor trained on data $\mathcal{D}$,
quantifies its average marginal contribution to all possible subsets of the dataset:
\begin{equation}
\phi_i=C \sum_{S \subseteq \mathcal{D}-\{i\}} \frac{V(S \cup\{i\})-V(S)}{\binom{n-1}{|S|}},
\end{equation}
where $V(S)$ is the performance metric on the training subset $S$ and $C$ is a constant.
Due to the computational complexity of calculating exact Shapley values, the authors propose approximation methods, including Monte Carlo sampling and Gradient-based approximations. \citet{wang2024data} addressed the computational challenges associated with traditional data Shapley calculations by introducing in-run data Shapley. The approach leverages the iterative nature of training algorithms, using first- and second-order Taylor expansions to approximate the impact of each data point on the model's performance. However, data Shapley values may mislead data valuation, especially in the presence of strong correlations among data points. \citet{wangrethinking} proposed refinements by grouping similar data points and computing Shapley values for clusters rather than individual points to account for redundancy.

In addition to calculating a static influential score, the temporal dependence of the training data quantifies the impact of omitting a particular data point at a specific training iteration \citep{wang2024capturing} by computing the dot product between the data point's embedding and the gradient of the test example's loss.

\subsubsection{Challenges and Future Directions}

The current state of data selection methods reflects a balance between efficiency, accuracy, and adaptability. While methods like Jaccard similarity offer simplicity, they lack semantic depth.  Jaccard similarity, a measure based on the size of the intersection divided by the size of the union of two sets, is often used to remove redundant data by identifying similar text samples. In the context of LLMs, it might be employed as a utility function to filter out duplicate or highly similar training examples, aiming to reduce dataset size without losing information. However, a key limitation is its inability to capture semantic meaning. For instance, two sentences with different wordings but similar meanings (e.g., ``The cat is on the mat'' vs. ``A feline rests atop a rug'') may have low Jaccard similarity in Equation \ref{eq:jaccard}, leading to their retention despite redundancy, or vice versa, potentially excluding valuable paraphrased data. This can result in suboptimal data selection, particularly for tasks requiring nuanced understanding, and highlights the need for semantic-aware measures.

Advanced techniques like LESS, STAFF, and Data Shapley address specific needs but are constrained by computational costs and practical limitations. Many methods, including LESS, STAFF, and data attribution methods like Data Shapley, involve training or evaluating models on subsets, which can be computationally expensive, especially for large datasets and models. For instance, STAFF's selection process can take over ten hours on powerful devices, driven by multiple epochs of training on the target model to evaluate data scores and regions, adding to computational cost.
LESS requires a preliminary training phase to obtain useful gradient features, increasing computational complexity and cost.
LLMs often involve massive datasets, and the dynamic nature of training means data importance can change over time, making static selection methods less effective. In-Run Data Shapley's \citep{wang2024data, wangrethinking} ability to capture training dynamics is a step forward, but adaptation remains a challenge.
To overcome the challenges, we suggest the following future research directions to enhance semantic understanding, reduce computational cost, and adapt to dynamic training dynamics, thereby fully leveraging the potential of LLMs. To save computational time and cost, we suggest using more efficient ways of data selection, subsampling based on the metric calculated by the original dataset \citep{meng2017effective, ma2022asymptotic, li2020modern, li2023core,wu2023subsampling} instead of metrics by training the model. Thus, we can reduce the computational cost by saving the model training time.

\section{Integration of Knowledge Distillation and Dataset Distillation}\label{sec:combine}

In this section, we examine the integration of KD and DD to mitigate the computational and scalability challenges inherent to modern LLMs. KD transfers nuanced reasoning skills, such as chain-of-thought capabilities, from large teacher models to compact students, while DD generates minimal yet representative datasets that preserve critical knowledge patterns. By combining these approaches, we demonstrate how their synergy reduces dependency on large-scale data, improves computational efficiency, and maintains advanced functionalities in distilled models. This integration establishes a cohesive framework for balancing model compression with sustainable data utilization.

\subsection{Knowledge Transfer via Dataset Distillation}\label{Jing Zhang,soheil kolouri,Ali Abbasi} 

\begin{figure}[h]
\centering
\includegraphics[width=0.9\linewidth]{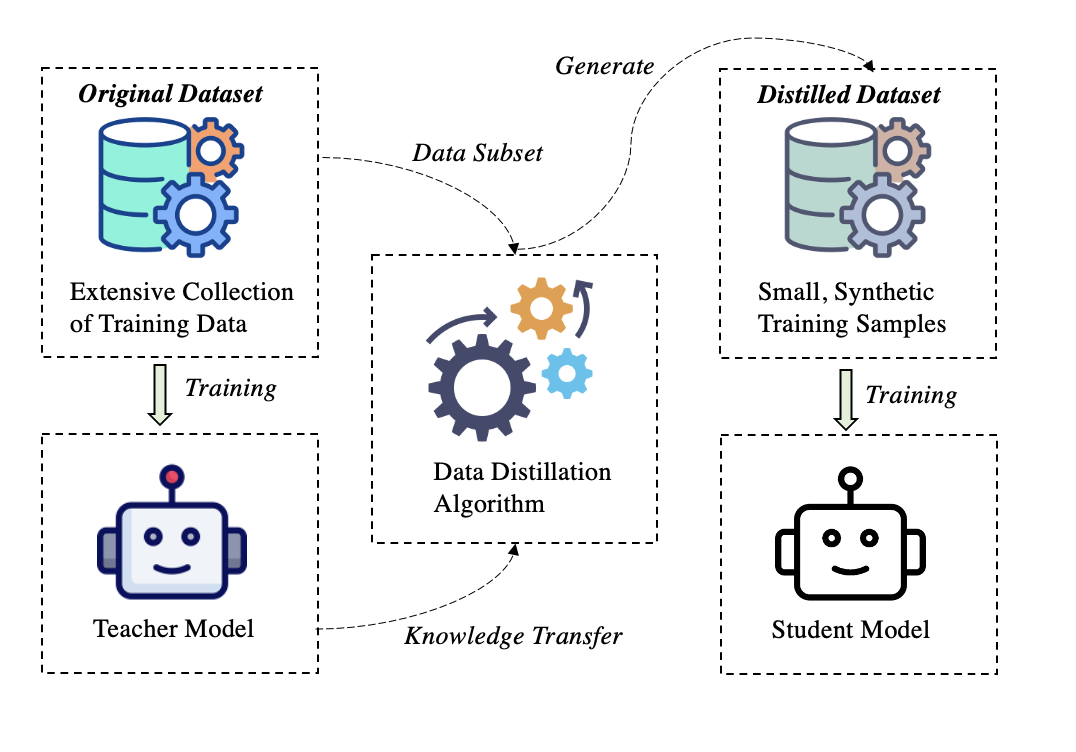}

\caption{An illustration of Knowledge Transfer via Data Distillation. The process illustrates how a teacher model trained on the original large dataset transfers its knowledge through distillation processes to create a small, synthetic dataset. This distilled dataset enables the efficient training of student models while maintaining performance.} \label{fig:data_distil}

\end{figure}

While meta-learning and bi-level optimization-based dataset distillation have demonstrated effectiveness across various small-scale benchmarks, they often suffer from two critical issues: (1) overfitting to the training dynamics during the distillation phase, and (2) limited scalability as the dataset size or model architecture complexity increases. This is primarily because optimization-based distillation methods rely on unrolling and storing the computation graph of multiple gradient steps during the distillation phase, an approach that becomes increasingly prohibitive in modern experimental settings. Moreover, their dependence on specific gradient steps leads to overfitting to the teacher architecture.

Recent research, building on these insights while addressing prior limitations, has developed integrated knowledge transfer frameworks that combine KD and DD techniques. Figure \ref{fig:data_distil} illustrates this unified methodology. More specifically, \citet{yin2023squeeze} recently proposed SRe2L, a method designed to disentangle meta-learning and bi-level optimization by reframing dataset distillation as data-frugal knowledge distillation. SRe2L follows a three-stage process: (1) pretraining the teacher model on a large-scale dataset, (2) leveraging model inversion \citep{yin2020dreaming} to synthesize a coreset, and (3) transferring the teacher's soft labels and the coreset to the student model. The objective of SRe2L is to learn a compact synthetic dataset \( C_{\text{syn}} \), composed of synthetic input-label pairs \( (\tilde{\boldsymbol{x}}, \tilde{y}) \), that encapsulates essential information from the original dataset \( \mathcal{D} \), which consists of real samples \( (\boldsymbol{x}, y) \). The learning process involves training a neural network \( \varphi_\theta \) on the synthetic data to minimize the expected classification loss:
\begin{equation}
\theta_{C_{\text{syn}}} = \arg\min_{\theta} L_C(\theta), \quad \text{where} \quad L_C(\theta) = \mathbb{E}_{(\tilde{\boldsymbol{x}}, \tilde{y}) \in C_{\text{syn}}} \left[ \ell(\varphi_{\theta_{C_{\text{syn}}}}(\tilde{\boldsymbol{x}}), \tilde{y}) \right],
\end{equation}
where \( \ell(\cdot) \) denotes the training loss function, implemented as the soft-label cross-entropy loss using teacher outputs obtained during the relabel stage:
\begin{equation}
\ell(\varphi_\theta(\tilde{\boldsymbol{x}}), \tilde{y}) = -\tilde{y} \cdot \log(\varphi_\theta(\tilde{\boldsymbol{x}})),
\end{equation}
where \( \tilde{y} \) is the soft label predicted by a teacher network trained on the original dataset.

To ensure that the synthetic dataset induces model training dynamics similar to those of the original data, SRe2L further constrains the optimization by minimizing the worst-case generalization gap:
\begin{equation}
\sup_{(\boldsymbol{x}, y) \sim \mathcal{D}} \left| \ell(\varphi_{\theta_T}(\boldsymbol{x}), y) - \ell(\varphi_{\theta_{C_{\text{syn}}}}(\boldsymbol{x}), y) \right| \leq \epsilon,
\end{equation}
where \( \theta_T \) are the parameters of a model trained on the full dataset \( \mathcal{D} \), and \( \epsilon \) is a small bound on the acceptable discrepancy in generalization performance between the two models. By decoupling dataset optimization and neural network training, SRe2L reduces computational costs while improving scalability. These synthesized samples, along with the soft labels, are then used to perform knowledge distillation on the student model.

Following SRe2L, several knowledge distillation-based dataset distillation approaches have emerged. \citet{shao2024generalized} introduced a method that employs diverse architectures during the distillation phase to enhance generalizability between architectures. \citet{sun2024diversity} proposed an approach that prunes image pixels by selecting the most informative patches and assembling them into a collage. Additionally, \citet{yin2023dataset} introduced advanced cropping techniques to further refine the quality of distilled samples. Moreover, although knowledge-distillation-based dataset distillation has proven highly effective in mitigating the computational complexity of dataset distillation, challenges remain, particularly in storing and communicating data and soft-label correspondence for training downstream students \citep{xiao2024large}.

Overall, knowledge transfer via data distillation represents a promising direction that bridges dataset distillation and knowledge distillation paradigms. By reformulating the problem as transferring knowledge from teacher to student through synthetic data, these methods effectively address the computational bottlenecks of traditional optimization-based approaches while maintaining competitive performance. This integration of concepts offers a more scalable and adaptable framework for knowledge compression that continues to evolve as researchers develop more sophisticated techniques for sample synthesis and information preservation.

\subsection{Prompt-Based Synthetic Data Generation for LLM Distillation}\label{Jiazhang}

\begin{figure}[!ht]
\centering
\vspace{-10pt}
\includegraphics[width=0.9\textwidth]{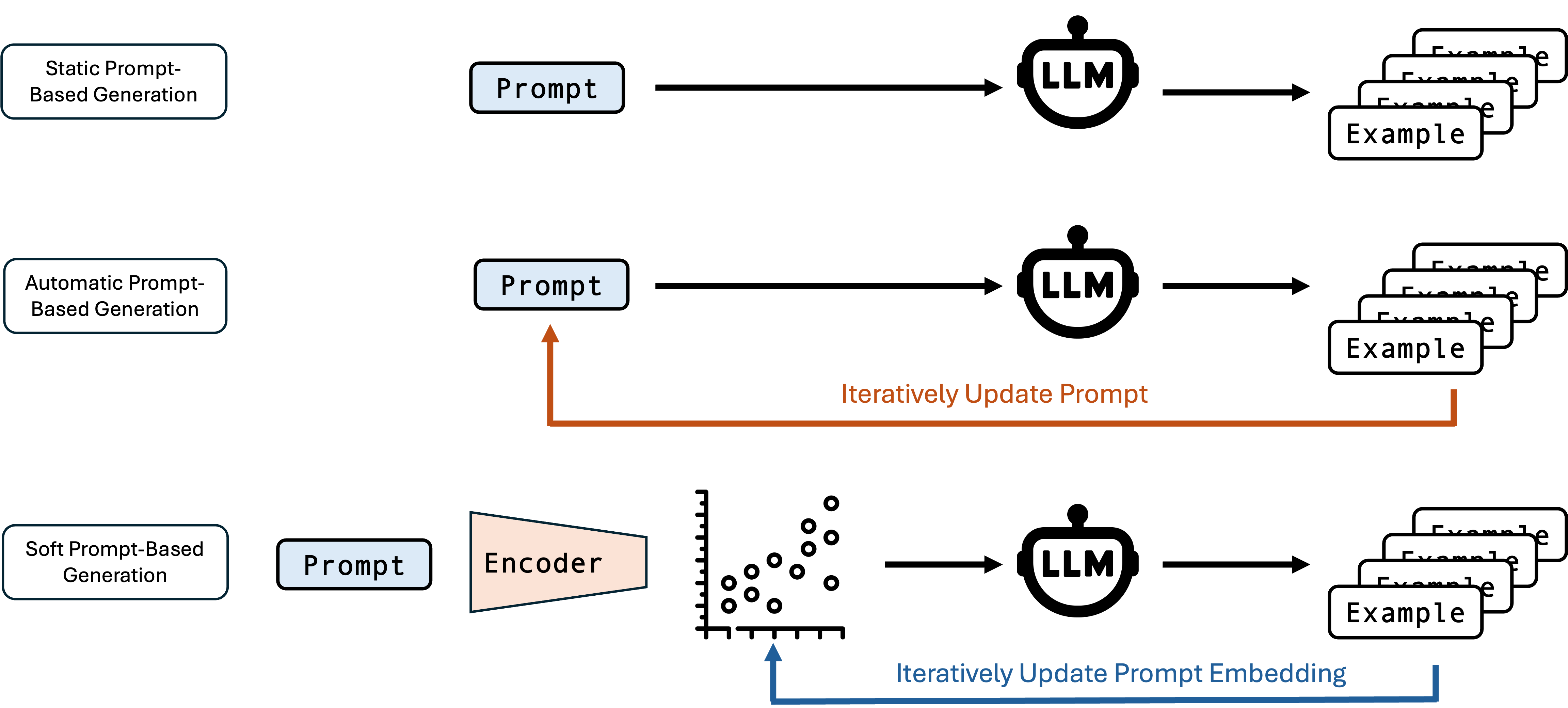}
\caption{Illustrations of Different Types of Prompt-based Synthetic Data Generation. All three types of methods generate task-related samples by providing a prompt to a well-trained LLM. Static prompt-based generation directly inputs the prompt into the target LLM. Automatic prompt-based generation iteratively updated the input prompt to get more complete and diverse samples. Soft prompt-based generation encodes the prompt into an embedding space and iteratively updates the embedding of the prompt, which converts the discrete prompt into a continuous value.}\label{fig:prompt_based}
\end{figure}

Another emerging trend is prompt-based synthetic data generation, which synergistically integrates KD and DD. Here, large-scale teacher LLMs generate compact, high-quality training sets through strategically designed prompts, transferring their knowledge into synthetic data while enabling efficient student model training. This unified approach has proven effective for low-resource adaptation, task-specific augmentation, and self-distillation.
The methodologies can be categorized into three main types: static prompt-based generation, automatic prompt optimization, and soft prompt-based frameworks.

Static Prompt-Based Generation involves using fixed, manually crafted prompts to direct LLMs in producing synthetic data. Researchers design prompts that elicit desired responses from the model, generating data that aligns with specific tasks or domains.
For example, PromDA \citep{wang2022promda} and GAL \citep{he2022generate} use manually crafted prompts to augment data for natural language understanding tasks in low-resource settings.

Automatic Prompt Optimization refers to techniques that iteratively refine prompts to enhance the quality and relevance of the generated data. 
This process involves using algorithms to adjust prompts based on feedback from the LLM's outputs, aiming to maximize certain performance metrics. 
For instance, \citet{deng2022rlprompt} introduces a reinforcement learning approach to automatically optimize discrete text prompts for language models. 
\citet{pryzant2023automatic} presents ProTeGi, a method for automatic prompt optimization using gradient-based techniques and beam search strategies.

Soft Prompt-Based Frameworks utilize learnable embeddings, known as soft prompts, to steer LLMs without altering their internal parameters. 
Unlike traditional text-based prompts, soft prompts are continuous vectors optimized during training to induce the desired model behavior. 
This method enables more nuanced control over the generated data and can be particularly effective in producing structured or domain-specific content. 
For example, DiffLM \citep{zhou2024difflm} introduces a controllable data synthesis framework that leverages diffusion language models to enhance the quality of synthetic data generation, particularly for structured formatted data like tabular and code data.
SoftSRV framework \citep{desalvo2024no} introduces a novel approach by employing soft prompts - trainable vectors - to steer frozen pre-trained LLMs toward generating targeted synthetic text sequences. 
This method allows for the creation of domain-specific synthetic data without extensive manual prompt engineering, enhancing the adaptability and efficiency of synthetic data generation.

These methodologies highlight the versatility of prompt-based data generation in leveraging LLMs to create synthetic datasets tailored to specific needs. 
By employing static prompts, automatic optimization, or soft prompt-based frameworks, researchers can effectively generate data that enhances model training and performance across various applications.

\section{Evaluation and Metrics for LLM Distillation Techniques}\label{sec:eval}

\subsection{Evaluation}\label{Yang Zhenyuan}

Evaluating distillation techniques for LLMs demands a rigorous framework that measures performance, efficiency, robustness, and knowledge transfer efficacy. This ensures distilled models are systematically assessed on their ability to retain task effectiveness while balancing computational and data efficiency.  

{\color{blue}To provide concrete reference points for evaluating distillation methods, we incorporate widely-used standardized benchmarks such as GLUE \citep{wang2021adversarial} for natural language understanding tasks, GSM8K \citep{cobbe2021training} and MATH \citep{hendrycks2021measuring} for mathematical reasoning evaluation, and MMLU \citep{wilkins2024higher} for comprehensive knowledge assessment. These benchmarks enable meaningful comparison across different distillation approaches and provide standardized evaluation protocols. Typical performance retention rates show student models achieving 90-95\% of teacher performance on these benchmarks while reducing model size by 5-10x, demonstrating the practical trade-offs between performance and efficiency in the field.}

\paragraph{Performance Metrics.} 
A key approach to evaluating distilled models is to assess their performance, with a primary focus on accuracy and generalization. Common evaluation metrics include perplexity for language modeling, exact match (EM) and F1-score for question answering, and BLEU or ROUGE scores for text generation. These metrics aim to quantify how effectively the student model preserves the predictive capabilities of the teacher model while minimizing performance degradation. Additionally, similarity measures such as cosine similarity and KL divergence between the teacher’s and student’s outputs provide insight into the extent of knowledge transfer.

Recent research in the LLM literature has introduced several evaluation metrics specifically designed for large language models. One such metric is the MAUVE score, which quantifies the divergence between the probability distributions of human-generated and model-generated text \citep{pillutla2021mauve}. It is formulated based on the Jensen–Shannon divergence, providing a principled measure of distributional similarity.
\begin{equation}
    \text{MAUVE} = \exp\left(-D_{\text{JS}}\left(P_{\text{human}} \parallel P_{\text{model}}\right)\right),
\end{equation}
where \(D_{\text{JS}}\) denotes the Jensen-Shannon divergence between the distribution \(P_{\text{human}}\) of human texts and \(P_{\text{model}}\) of model outputs. 

Another metric, BERTScore, measures the similarity between generated and reference texts by computing the cosine similarity between contextual embeddings \citep{zhang2019bertscore}. Its formulation can be written as
\begin{equation}
    \text{BERTScore} = \frac{1}{T} \sum_{t=1}^{T} \max_{s \in S} \cos\left(\mathbf{e}_t, \mathbf{e}_s\right),
\end{equation}
where \(\mathbf{e}_t\) and \(\mathbf{e}_s\) represent the embedding vectors of tokens in the candidate and reference sentences respectively, \(T\) is the number of tokens in the candidate text, and \(S\) is the set of tokens in the reference text. 
Other metrics, such as Self-BLEU, which quantifies repetition by computing the BLEU score of a generated text against itself, and distinct-$n$, which is defined as the number of unique $n$-grams normalized by the total number of $n$-grams, address critical aspects of text quality like diversity and consistency. Consistency metrics, measured as the agreement rate, compute the agreement rate of multiple model responses to similar prompts to ensure stability in open-ended generation tasks.

For higher-order capabilities, evaluation extends to specialized benchmarks. For example, reasoning ability is measured through mathematical problem-solving (e.g., GSM8K \citep{cobbe2021training}, MATH \citep{hendrycks2021measuring}) and logical deduction tasks \citep{liu2020logiqa}, while natural language generation is assessed via summarization fidelity, question answering accuracy, and retrieval-augmented generation performance in search engine contexts \citep{min2023factscore}. For a systematic taxonomy of LLM evaluation methodologies, we refer to the comprehensive survey by \citet{chang2024survey}.

\paragraph{Complexity and Efficiency} Distillation aims to reduce the computational footprint of LLMs. Evaluation in this aspect involves measuring inference speed, memory consumption, and model size. Metrics such as FLOPs (floating point operations), latency, and peak memory usage provide quantitative benchmarks for comparing different distillation techniques. {\color{blue}Specific efficiency metrics include inference latency measurements showing reductions from hundreds of milliseconds to tens of milliseconds per query, and memory consumption comparisons demonstrating decreases from gigabytes to hundreds of megabytes for model storage and inference.}  Efficiency gains are particularly relevant for edge and low-resource deployment scenarios where computational constraints are stringent.

\paragraph{Robustness and Uncertainty.} 
Ensuring the robustness of distilled models under adversarial conditions and distribution shifts is critical for reliable deployment. Robustness evaluation involves stress-testing models against domain shifts, adversarial perturbations, and noisy inputs. The Attack Success Rate (ASR) measures the efficacy of adversarial attacks in fooling the model \citep{wang2023robustness,wang2021adversarial}. For a dataset $\mathcal{D}=\left\{\left(x_i, y_i\right)\right\}_{i=1}^N$ containing $N$ input-output pairs and an attack method $\mathcal{A}$ that generates adversarial examples $\mathcal{A}(x)$, ASR is defined as:
\begin{equation}
    A S R=\sum_{(x, y \in D)} \frac{I[f(\mathcal{A}(x)) \neq y]}{I[f(x)=y]} ,
\end{equation}
where $I$ is the indicator function, $f$ is the model under test, and the denominator counts samples correctly classified before the attack.
To evaluate prompt-based robustness, the Performance Drop Rate (PDR) quantifies relative degradation after adversarial modifications to prompts \citep{zhu2023promptbench}. For an original prompt $P$, adversarial prompt $A(P)$, and task-specific evaluation metric $\mathcal{M}$ (e.g., accuracy or F1-score), PDR is calculated as:
\begin{equation}
    \operatorname{PDR}=1-\frac{\sum_{(x, y) \in \mathcal{D}} \mathcal{M}[f([A(P), x], y)]}{\sum_{(x, y) \in \mathcal{D}} \mathcal{M}[f([P, x], y)]}
\end{equation}
where higher PDR indicates greater vulnerability to prompt attacks.

Calibration and uncertainty estimation ensure reliable confidence scores. The Expected Calibration Error (ECE) partitions model predictions into $M$ equally spaced confidence bins $B_1, \ldots, B_M$ and measures the discrepancy between accuracy and confidence within each bin:
\begin{equation}
    \mathrm{ECE}=\sum_{m=1}^M \frac{\left|B_m\right|}{N}\left|\operatorname{acc}\left(B_m\right)-\operatorname{conf}\left(B_m\right)\right|,
\end{equation}
where $\left|B_m\right|$ is the number of samples in bin $m, \operatorname{acc}\left(B_m\right)$ is the bin's accuracy, and $\operatorname{conf}\left(B_m\right)$ is the average predicted confidence \citep{guo2017calibration,tian2023just}. 
Entropy-based uncertainty estimates prediction stability using Uncertainty $=-\sum_y p(y \mid x) \log p(y \mid x)$, where higher entropy reflects lower confidence in outputs.

\subsection{Quantifying Distillation Level in LLMs} \label{Yang Zhenyuan}

A critical challenge in LLMs development lies in quantifying how much knowledge is distilled from teacher models, particularly since unregulated distillation risks model homogenization, identity leakage, and robustness degradation.

One approach is to compute the divergence between the probability distributions of teacher and student models using KL divergence or Jensen-Shannon divergence. These metrics help in determining the extent of information retained post-distillation.
Another method involves evaluating feature representations extracted from different layers of the teacher and student models. Cosine similarity and centered kernel alignment provide a way to compare internal representations, offering insights into structural similarities between models. Additionally, task-specific evaluation can reveal how distillation affects downstream performance.

Recent work \citep{lee2025distillation} advances distillation quantification through Identity Consistency Evaluation (ICE) and Response Similarity Evaluation (RSE). These methods systematically measure identity leakage (e.g., Qwen-Max-0919 falsely attributing its development to Claude in 32\% of cases) and output homogenization (e.g., DeepSeek-V3 achieving RSE $>4.1$ against GPT-4o) to assess distillation efficacy. These methods highlight critical trade-offs: while distillation improves efficiency, it risks robustness degradation and reduced model diversity.

\section{Applications and Use Cases of Distillation}\label{sec:apply}

This section surveys knowledge distillation techniques across specialized domains, including medical and healthcare, education, and bioinformatics, demonstrating their transformative impact in optimizing domain-specific AI systems.

\subsection{Medical and Healthcare}\label{Zhengliang Liu}
Recent advancements in KD for LLMs have enabled more efficient and specialized applications in healthcare. Distillation techniques allow for the compression of large models into smaller, task-specific ones while preserving essential capabilities, making them practical for clinical deployment. Below, we highlight recent research contributions in clinical decision support, medical summarization, patient interaction, and drug discovery.

\subsubsection{Clinical Decision Support}

ClinRaGen \citep{niu2024clinragen} exemplifies how KD can be harnessed to build efficient clinical decision support systems by equipping small language models (SLMs) with LLM-level reasoning. The framework first retrieves disease-specific medical knowledge and uses an LLM (e.g., ChatGPT) to generate structured rationales from multimodal EHR data that combine textual clinical notes and time-series lab results and serve as high-quality distillation targets. Through a three-phase sequential distillation process, including medical note–based rationale learning, knowledge-augmented attention for lab test rationale, and full multimodal integration, the student SLM is distilled into a compact 80M‑parameter model, which is more than 2,000x smaller than the 175B‑parameter teacher LLM and 80x smaller than a fine‑tuned 7B‑parameter LLaMA model, yet it achieves excellent diagnostic performance on MIMIC‑III and MIMIC‑IV while training in under half the time. In rationale quality evaluations with GPT‑4 and human judges, ClinRaGen ranks second for readability and correctness—matching the larger LLaMA3 model—and outperforms other 7B‑ and 60M‑parameter models in consistency and clinical soundness. A larger variant, ClinRaGen* (793M parameters), further boosts F1 by over 1.5\%, highlighting the favorable trade‑off between scale and performance.



\citet{ding2024ckle} proposes CKLE, a framework for ICU health event prediction through KD from general LLMs into multimodal electronic health record (EHR) models. Their approach transfers knowledge from a text-based teacher LLM to a smaller student model that processes both clinical text and structured patient data using cross-modal contrastive objectives. The distilled model improved prediction accuracy for heart failure and hypertension by up to 4.48\% over state-of-the-art models while addressing privacy and deployment constraints through local, smaller models with LLM-level insights. The significance of this approach is underscored by similar real-world deployments at major medical centers, such as Mount Sinai's Advanced Alert Monitor program, which has demonstrated that AI-based prediction systems can save more than 500 lives per year when properly integrated into clinical workflows.

\citet{hasan2024optimclm} introduces OptimCLM, a comprehensive compression framework combining ensemble learning, knowledge distillation, pruning, and quantization for clinical BERT models. Their method uses an ensemble of domain-specific BERTs (DischargeBERT and COReBERT) to teach compact student models (TinyBERT and BERT-PKD) via black-box distillation. For hospital outcome tasks including length of stay, mortality, diagnosis, and procedure prediction, the student models achieved up to 22.8× model size compression and 28.7× speedup with minimal performance loss (under 5\% decrease in AUROC), demonstrating that knowledge distillation can preserve accuracy while dramatically reducing computational requirements.

\subsubsection{Medical Summarization}


\citet{tariq2024radiology} presents a novel application of knowledge distillation to develop a patient‑centric radiology report summarization system. They leverage a 13B-parameter LLaMA model as a teacher to generate noisy layman summaries for approximately 7K chest CT reports, and then fine‑tune a 770M-parameter T5 student model on this weakly labeled data, reducing model size by over 17× while maintaining high fidelity. Compared to LLaMA zero‑shot outputs, the distilled student cuts hallucinations from 18\% to 6\% and missing information from 17\% to 4\%. Expert radiologists and lay users rate the student’s summaries as factually accurate and substantially more understandable, improving patient comprehension by 63\%, all with far lower computational cost, demonstrating how distillation can yield efficient, reliable clinical decision support tools. 

While explicit applications of KD for medical summarization are still emerging, large LLMs such as GPT-4 have shown impressive ability in this domain, in some cases producing summaries non-inferior to physicians. The success of LLMs in summarizing clinical text suggests significant opportunities to distill these capabilities into lightweight summarizers for electronic health records, representing a promising area for future research.

\subsubsection{Patient Interaction (Q\&A and Matching)}

\citet{nievas2024clinicaltrials} presents Trial‑LLAMA, an open‑source LLAMA model fine‑tuned for patient–trial matching. They generate a synthetic dataset of 2,000 patient–trial reasoning examples using GPT‑4 and then supervise‑fine‑tune the LLAMA‑2‑70B model on this data. The resulting Trial‑LLAMA model matches or surpasses proprietary GPT‑3.5 baselines across key aggregate metrics such as NDCG, Precision, and AUROC, and achieves competitive criterion‑level accuracy in both explicit and implicit eligibility judgments. By relying on synthetic data and open‑source architectures, this work demonstrates a cost‑effective, privacy‑conscious approach to deploying large‑model reasoning in clinical decision support.

\citet{sutanto2024llmqa} addresses few-shot medical question answering by distilling knowledge from large LLMs into smaller models. Their approach uses a teacher LLM to generate synthetic multiple-choice questions with answer likelihood scores, creating a rich training set for a DeBERTa-v3 student model. On the MMLU benchmark, which includes medical exam questions, the distilled student achieved 39.3\% accuracy compared to 28.9\% for a baseline trained on few-shot data—a significant 10\% absolute improvement. This demonstrates how LLM-generated data and knowledge distillation can substantially enhance a compact model's medical reasoning capabilities.

\citet{yagnik2024medlm} compares general versus medical-specific distilled language models for open-ended medical question answering. Their evaluation indicates that medical-domain distilled models, such as BioMedBERT variants, can rival larger general models on expert question answering, reinforcing the value of distilling domain knowledge into compact chatbots for health-related queries.

\subsubsection{Clinical Information Extraction and Knowledge Curation}
\citet{gu2023ade} leverages GPT-3.5 and GPT-4 as teachers to train a compact biomedical model for extracting adverse drug events (ADEs) from text. Instead of relying on costly manual labels, they prompt GPT-3.5 to structure raw biomedical text, then distill its outputs into a PubMedBERT student via self-supervised learning. Remarkably, the distilled PubMedBERT (with 1,000× fewer parameters) outperformed its teacher: achieving 6 points higher F1 than GPT-3.5 and even 5 points higher than GPT-4 on standard ADE extraction. Similar gains were shown for other information extraction tasks, demonstrating that distillation can transfer LLM capabilities into small, domain-specific models that are both efficient and high-performing.

\citet{vedula2024cie} explores using LLMs as labelers for clinical text, distilling their knowledge into BERT-based models for clinical named entity recognition tasks. By generating pseudo-labels on clinical notes using GPT-4-level models and medical ontologies as teachers, they trained a BioBERT student that achieved F1 scores close to a fully supervised model while running 12× faster and at 1\% of the cost of the large LLM in inference. This demonstrates that distillation can deliver nearly the same accuracy as an LLM on clinical text extraction while being dramatically more efficient in speed and cost—enabling practical deployment in hospital data pipelines.

\subsubsection{Drug Discovery}
\citet{zhang2025kedrec} introduced KEDRec-LM, an instruction-tuned LLM for explainable drug recommendation. The model leverages knowledge from biomedical literature, drug repurposing databases, and clinical trials. The authors use a teacher LLM to generate domain-specific rationales, which are then distilled into a student model. The distilled model effectively synthesizes biomedical knowledge to support drug–disease relationship analysis and hypothesis generation.

By constructing a large training corpus integrating a drug repurposing knowledge graph, clinical trial data, and PubMed articles, then distilling this multi-source knowledge into a generative LLM, KEDRec-LM achieves state-of-the-art performance on drug selection tasks. On the expRxRec dataset, it achieved F1 scores up to 88.05\%.

\subsubsection{Multimodal and Vision-Language Applications}
\citet{ge2025clinkd} introduces ClinKD, a framework to boost medical visual question answering (Med-VQA) performance by injecting prior knowledge via distillation. The model builds on a vision-language architecture capable of analyzing both images and text questions. A teacher model incorporating Med-CLIP visual features and a large multimodal transformer is used to teach the student model about image-text alignment and medical concepts through a distillation phase. ClinKD achieved state-of-the-art accuracy on the Med-GRIT 270k benchmark for medical image QA, demonstrating superior ability to answer questions about radiology images, including pinpointing anatomical regions. By distilling cross-modal knowledge, the approach addresses issues like visual hallucination and improves the grounding of answers in medical images, highlighting how vision-language LLMs in healthcare can be made more compact and accurate through targeted knowledge distillation.

These studies collectively illustrate the transformative role of distillation techniques in adapting LLMs for healthcare applications. By reducing computational costs while maintaining high performance, knowledge-distilled models hold promise for scalable and interpretable AI-driven medical solutions that can be deployed in real-world clinical settings.

\subsubsection{Key Lessons Learned}

Distillation enables compact clinical models to rival large LLMs. Previous examples show that even with tens, hundreds, or even thousands of times fewer parameters, student models can still achieve comparable results using distillation techniques, which makes AI more deployable in resource-constrained clinical settings by reducing computational costs while preserving diagnostic accuracy. However, distillation in healthcare must contend with fragmented, imbalanced datasets that can undermine student robustness, and biases inherent in teachers or synthetic examples can propagate and even amplify disparities in underrepresented patient groups \citep{cross2024bias}. Federated distillation offers a privacy‑preserving alternative by exchanging distilled outputs rather than raw records, but it faces challenges with heterogeneous site distributions and potential information leakage \citep{salman2025knowledge}. Moreover, the memory and compute demands of loading large teachers during offline distillation often exceed the capacities of typical hospital infrastructure \citep{templin2024addressing}.

\subsection{Education}\label{Ehsan Latif/Xiaoming Zhai}
LLMs have demonstrated remarkable capabilities in automated scoring, lesson planning, and other generative tasks in education \citep{zhai2022chatgpt,lee2024using}. However, their practical deployment in real-time educational settings faces significant challenges, particularly when considering resource-constrained environments such as schools with low-end devices, mobile tablets, and laptops with limited computational power \citep{selwyn2019should,hinton2015distilling}. The primary obstacles include:

\begin{itemize}
    \item \textbf{High Computational Requirements}: LLMs such as BERT-based models require significant processing power, often necessitating specialized hardware such as GPUs or TPUs. The necessity for such resources limits the accessibility of these models in standard school environments, where computational resources are minimal \citep{hinton2015distilling}.
    \item \textbf{Large Model Size}: The memory footprint of LLMs, often in the range of hundreds of megabytes, presents difficulties in deploying them on edge devices or cloud-limited school networks. \citet{latif2024knowledge} report that their teacher LLM has approximately 114 million parameters, making direct deployment infeasible on low-resource devices.
    \item \textbf{Inference Latency}: The complexity of LLMs results in longer processing times, making real-time feedback and scoring challenging \citep{moon2010online}. Latency is a critical factor in automated assessment, where immediate results are needed for adaptive learning and student engagement.
\end{itemize}

To address these challenges, researchers have explored model compression techniques such as KD to create smaller yet efficient models for educational assessments \citep {hinton2015distilling}. The goal is to retain the accuracy and predictive capabilities of the teacher model while significantly reducing computational overhead. This is achieved by training the student model using the soft labels generated by the teacher model instead of hard class labels. \citet{latif2024knowledge} proposes a KD-based approach specifically designed for educational assessments. Their methodology involves the utilization of a fine-tuned BERT model as the teacher model for scoring student responses in science and mathematical reasoning assessments. Their study demonstrates that the KD approach achieves a drastic reduction in model size; for example, the student model is 4,000 times smaller than the teacher model, making it feasible for deployment on low-end educational devices. The method has reduced inference time, such as the KD model, which is 10 times faster in inference compared to the teacher model, ensuring real-time scoring capabilities. Finally, the high scoring accuracy achieved by the distilled student model with a mere 2-3\% accuracy reduction compared to the teacher model, while outperforming existing state-of-the-art distilled models such as TinyBERT \citep{jiao2019tinybert}.

Other studies have also explored the distillation of LLMs specifically tailored for educational contexts. For instance, \citet{dan2023educhat} introduced EduChat, a large-scale LLM-based chatbot designed to facilitate intelligent education. EduChat utilizes KD to improve conversational efficiency, providing scalable and personalized educational interactions. Similarly, \citet{qu2024coursegpt} developed CourseGPT-ZH, an educational LLM incorporating KD and prompt optimization, significantly enhancing performance in pedagogical settings by efficiently distilling domain-specific knowledge into more manageable model sizes. Additionally, \citet{baladon2023retuyt} fine-tuned open-source LLMs specifically for generating teacher-like responses, underscoring the practical applications of KD in real-world educational scenarios.

Another representative example is EduChat, developed by \citet{dan2023educhat}, which demonstrates large-scale real-world deployment of distilled LLMs in education. EduChat is a conversational AI platform tailored for use in Chinese primary and secondary schools, supporting subjects such as Chinese, mathematics, English, physics, chemistry, and history. In its initial deployment, EduChat was piloted across over 200 schools, serving a student population of more than 50,000. The developers applied KD to shrink their base LLM by about 80 percent compared with the teacher model. Memory use dropped from more than 40 GB to less than 8 GB, allowing the chatbot to run smoothly on standard school servers and many high-end tablets. This size reduction allowed EduChat to deliver response times averaging under 1.5 seconds per query, even during peak classroom use, and to support thousands of simultaneous student sessions without performance degradation. In practice, the distilled EduChat model was able to provide real-time, curriculum-aligned explanations, quiz generation, and homework help across a variety of subjects, greatly expanding the reach of AI-powered tutoring in resource-constrained environments. However, field deployment also highlighted certain limitations: while the distilled model handled factual and procedural questions with high reliability, it sometimes struggled with more open-ended, creative, or cross-disciplinary prompts. This finding underscores the ongoing challenge of transferring deep reasoning abilities through distillation.

Beyond direct educational assessment, KD approaches have demonstrated potential in facilitating a deeper understanding of scientific concepts, which can significantly benefit education. The Darwin series, introduced by \citet{xie2023darwin,xie2024darwin}, developed domain-specific LLMs tailored for natural sciences and materials science education, respectively. These models leverage KD to optimize domain-specific knowledge representation, improving accessibility and comprehension in educational contexts. \citet{dagdelen2024structured} further advanced this idea by leveraging structured information extraction from scientific texts through LLMs, illustrating how sophisticated knowledge distillation techniques can enhance educational content delivery and learner engagement.

The success of knowledge distillation in automated educational assessment paves the way for scalable AI adoption in schools. By leveraging KD, LLM-powered assessment systems can be deployed on widely available hardware, reducing reliance on expensive cloud-based solutions. This democratizes AI-driven education, making advanced assessment tools accessible to a broader range of learners and institutions \citep{selwyn2019should}. Future research can further enhance KD strategies by incorporating adaptive loss functions, multimodal distillation techniques, and task-specific optimizations to improve performance across diverse educational applications \citep{fang2025efficient}. By refining KD methodologies, educational AI can become more practical, efficient, and equitable in real-world deployment scenarios.

\textbf{Key Lessons Learned.}
Across these cases, several lessons emerge. First, KD enables the deployment of high-performing educational AI on common hardware by drastically reducing model size and inference time, democratizing access to advanced tools. Second, there is often a small but measurable drop in performance, which may affect particularly complex or nuanced tasks. This makes the choice of what to distill, and how, especially significant. Third, model distillation is not just about efficiency: by making LLMs available in more settings, it can transform how feedback and assessment are delivered, making real-time, personalized education feasible at scale. Future work should focus on minimizing knowledge loss in distillation and optimizing models for diverse educational needs, including subject-specific tasks and multimodal learning contexts \citep{fang2025efficient}.

\subsection{Bioinformatics}\label{Jiazhang, Yufang}

Knowledge Distillation (KD) in bioinformatics serves as a pivotal strategy to compress large neural architectures into efficient models tailored for specific biological tasks. 
Broadly, these applications fit into four practical domains: protein representation learning, genomic regulatory modeling, drug-target affinity prediction, and biomedical named entity recognition.
Below, we present a cohesive overview organized by usage, highlighting multiple studies per category to illustrate both scope and depth.

\subsubsection{Protein Representation Learning}

In protein representation learning, knowledge distillation tackles the scalability challenge posed by large protein language models (PLMs). 
The typical pipeline starts with selecting several high-capacity PLMs, such as ESM, ProtTrans, or ProtT5, as teacher models, which separately learn rich representations of protein sequences. 
A compact student network is then trained using multiple objectives: a soft-label loss to replicate teacher predictions, a feature-matching loss to align intermediate activations, and occasionally, meta-learning components that adaptively weight teacher contributions based on each protein's characteristics. 
For example, \citet{shang2024accurate} demonstrates that a distilled student maintained within ±1.5\% accuracy of the teacher ensemble while achieving ~70\% faster inference. 
\citet{wang2022contact} in Contact‑Distil translates ProtT5 embeddings into a smaller TCN-BiLSTM architecture to preserve contact map fidelity, particularly in data-scarce regimes.
\citet{zhang2023adaptive} introduces meta-teacher selection to dynamically influence the student based on task difficulty or protein family, boosting adaptability. 
Additionally, KD-MultiSucc \citep{tran2025kd_multisucc} synthesizes cross-species teachers for better post-translational site prediction, showing that even highly specialized embeddings benefit from multi-teacher fusion.

\subsubsection{Genomic Regulatory Modeling}

Regulatory genomics demands models that combine interpretability and scalability. 
Ensemble methods deliver high accuracy, but their computational cost is prohibitive. 
The distilled pipelines begin by training an ensemble of neural networks on regulatory elements—such as transcription factor binding sites or chromatin states—then distill this ensemble into a compact student by minimizing both mean-squared error for predictions and KL-divergence for ensemble uncertainty. 
In some designs, auxiliary uncertainty losses, like aleatoric variance, further enhance the quality of calibration. 
\citet{zhou2024uncertainty} used this strategy to retain ensemble-level confidence in a lightweight model suitable for genome-scale scans, reducing inference time by over 50\%.
It also included the aleatoric uncertainty estimation to the student, leading to better attribution maps and uncertainty allocation. 
These pipelines highlight both the efficiency gains and the interpretive clarity achievable via KD, albeit with caution—ensemble reduction occasionally dulls edge-case detection, underscoring the need for future work on preserving ensemble diversity during distillation.

\subsubsection{Drug–Target Affinity Prediction}

In computational drug discovery, KD facilitates rapid screening of vast compound libraries. 
The pipeline integrates diverse teacher models—such as graph neural networks trained on drug structures, sequence-based protein encoders, or descriptor-based predictors, and uses an attention, guided fusion loss to distill rich feature representations into a student model. 
The objective is calibrated to recover teacher outputs while enforcing compactness. 
\citet{lu2023improving} showed that a student model could emulate complex multi-modal teachers with negligible accuracy loss and 60–80\% faster inference. 
Follow-up models like FusionDTA \citep{yuan2022fusiondta} and ViDTA \citep{li2024vidta} introduced graph attention mechanisms and virtual nodes to better model drug–protein interactions during distillation. 
These pipelines highlight KD’s role in creating scalable, high-throughput affinity predictors, though they also reveal persistent gaps in out-of-distribution chemical generalization, motivating the need for adaptive teacher selection and domain-augmented distillation.

\subsubsection{Biomedical Named Entity Recognition (BioNER)}

BioNER systems benefit from KD through model compression and label noise mitigation. 
The typical pipeline starts with using a large pretrained teacher (e.g., BERT or BioBERT) to generate soft labels or to correct noisy annotations. 
A student model, often either a BiLSTM-CRF or a distilled TinyBERT, is then trained using a combination of standard cross-entropy and KD-guided losses. 
\citet{zhou2021improving} refined noisy labels via teacher predictions before distilling into a compact student, leading to better recall and faster inference. 
\citet{jiao2019tinybert} adapted TinyBERT for biomedical corpora, achieving transformer-level accuracies with a much smaller footprint. 
More recent methods like ProTeGi \citep{mehmood2023distilling} extended KD to train non-transformer student models for protein–gene relation extraction, enabling deployment on edge devices.

\subsubsection{Key Lessons Learned}

Across the bioinformatics cases reviewed, several key lessons emerge.
First, knowledge distillation enables the deployment of complex biological models, such as large protein language models or genomic ensembles, on standard hardware by significantly reducing model size and inference time. 
This facilitates broader access to high-performance tools, particularly in environments with limited computational resources. 
Second, while distillation often preserves overall accuracy, it can introduce subtle but important performance drops, especially in tasks involving rare entities, noisy data, or complex biological patterns. 
This highlights the importance of choosing appropriate teacher models and designing domain-aware distillation strategies. 
Third, KD is not only a tool for efficiency, but it can also reshape how biological data is processed and analyzed, enabling real-time sequence annotation, scalable genome analysis, and faster experimental feedback. 
Future research should prioritize minimizing information loss during distillation and tailoring models to diverse biological contexts, including low-resource species, noisy experimental conditions, and multi-omics integration.

\section{Open Challenges and Future Directions}\label{sec:challenge}

Despite significant advances in KD and DD, several core challenges, especially those that are unique to or significantly amplified in the context of LLMs, remain unresolved. These challenges present exciting opportunities for future research in this rapidly evolving field. In Sections \ref{sec:method_KD} and \ref{sec:method_DD}, we discussed the limitations and future directions for specific KD and DD approaches. Here, we summarize the major open questions and challenges, aiming to provide guidance for essential and promising directions in future research.

\paragraph{Preserve the deeper context and reasoning knowledge.}
Traditional distillation methods (e.g., logit or layer matching) often fail to retain emergent large-scale capabilities such as chain-of-thought reasoning and in-context learning \citep{hinton2015distilling, gou2021knowledge}. Excessive compression or selective sampling further disrupts long-range dependencies required for tasks like narrative coherence and coreference resolution \citep{li2024chulo,bender2021dangers}. These challenges are compounded by language’s inherent diversity across domains, dialects, and specialized fields \citep{gururangan2020don}. Without explicit safeguards, nuanced linguistic elements, including rare idioms, archaic phrases, or technical jargon, risk being discarded, diminishing the adaptability and domain-specific utility of distilled models compared to general-purpose LLMs \citep{wang2024comprehensive}.
Future research should prioritize methodologies that explicitly extract and preserve deeper contextual relationships and reasoning mechanisms, such as chain-of-thought dynamics and in-context learning signals \citep{hsieh2023distilling,feng2024keypoint}. These strategies must also integrate domain-aware frameworks to retain linguistic diversity, from common usage to specialized vocabulary, ensuring distilled models maintain the structural and contextual fidelity of their larger counterparts.

\paragraph{Distillation cost.}
The exponential growth of LLMs, now reaching trillions of parameters, has amplified the computational and logistical burden of knowledge distillation. Iterative updates to these models, whether through fine-tuning or new data additions, force repeated distillation cycles that demand significant hardware resources and training time \citep{xu2024survey}. At the same time, compressing massive datasets without sacrificing linguistic diversity and rare patterns remains intractable with traditional techniques, such as gradient matching or Jaccard similarity, which either scale poorly or compromise semantic fidelity \citep{zhao2021dataset,khan2024lshbloom}. Balancing efficiency with the retention of nuanced knowledge is further complicated by static, heuristic-driven approaches that cannot adapt to ever-evolving models and data. Such methods risk discarding critical context, specialized vocabulary, and domain-specific phenomena crucial for robust downstream performance \citep{zhou2023distillspec,wang2024model}.
Looking ahead, the path forward requires the development of adaptive, lightweight frameworks that unify semantic-aware data compression, such as embedding-guided subsampling, and modular knowledge extraction \citep{wang2024data, wangrethinking}. By selectively targeting critical reasoning layers and avoiding full retraining, these strategies can reduce redundant computations and achieve real-time alignment with evolving model parameters and data distributions. Advances in dynamic trajectory alignment and hybrid generative-optimization approaches offer promising avenues for bridging efficiency gaps. Ultimately, success in distillation efficiency will hinge on designing methods that account for the sheer scale of models and corpora while preserving the depth and diversity essential to building robust, generalizable student models.

\paragraph{Trustworthy distillation.}
The rapid advancement of LLMs has amplified concerns about trustworthiness, including bias amplification and fairness degradation during distillation \citep{rejeleene2024towards,augenstein2023factuality,liu2023trustworthy}. When distillation is applied, a student model may inadvertently inherit and even intensify harmful artifacts from teacher models, particularly when compression discards nuanced contextual safeguards. 
Massive training corpora can also embed sensitive or private information that persists through distillation, raising additional ethical and legal concerns \citep{zhou2024security,dong2024building,shokri2017membership}. Hallucination is another serious challenge, as smaller models may further increase the likelihood of fabricating unfounded responses \citep{huang2025survey}.
A promising avenue to address some of these concerns is uncertainty quantification. By incorporating confidence-aware mechanisms (e.g., BKD described in Section \ref{sec:method_KD}'s uncertainty-aware KD part \citep{fangbayesian2024}), student models can flag high-uncertainty responses, prompting users or downstream systems to apply additional checks or override questionable results. Existing studies have also attempted to solve the trustworthiness problem by learning concentrated or reliable knowledge \citep{yang2024rlcd,shum2024first}. Integrating these techniques with rigorous data governance frameworks, which emphasize real-time monitoring and ethical oversight, can ensure distilled models maintain accuracy, fairness, and security as LLMs evolve.

\paragraph{Fairness and bias amplification in distillation.}

The process of KD and DD can inadvertently amplify biases present in teacher models and training data, creating significant ethical and practical concerns for deployed systems. This challenge manifests through several distinct mechanisms that require systematic understanding and mitigation strategies. 
In KD, the students can amplify biased patterns as compression optimizes for aggregate rather than individual fairness. Distillation losses can incentivize shortcuts through demographic correlations while sacrificing teacher models' subtle protection, developing less complex, more biased decision boundaries \citep{augenstein2024factuality,li2023survey}. This issue is more severe in self-distillation, as it is trained through feedback cycles in which biased outputs train the next iteration, continually amplifying the bias \citep{arazo2020pseudo,allen2020towards}.
In DD, data selection or synthesis regularly underrepresents minority groups. Classic measures such as gradient matching favor high-frequency patterns at the expense of the rare instances essential for equitable representation, generating datasets lacking minority instances along with required linguistic diversity for equitable behavior.

Overcoming these challenges entails fairness-aware distillation processes under the constraints of demographic parity or equalized odds. 
A key open question is how to set learning objectives that collectively maximize efficiency, fairness, and predictive performance without degenerating into insignificant solutions. Some promising directions involve multi-objective optimization strategies that actively control the balance between parameter compression and group-wise error gaps.
Additionally, current evaluations typically focus on aggregate metrics while neglecting fairness assessments. Future research should develop comprehensive protocols measuring bias amplification and fairness degradation across demographic groups, including specialized benchmarks and monitoring systems for distilled models.
There is a need to formulate sound distillation methodologies that balance fairness as they achieve efficiency gains because these frameworks operate in high-stake environments, which affect human well-being.

\paragraph{Dynamic evolution of teacher models and training corpora.}
The rapid evolution of LLMs introduces two intertwined challenges for distillation: the dynamic nature of teacher models (KD) and the continuous expansion of training corpora (DD). 
First, teacher models are frequently updated through methods such as fine-tuning, domain adaptation, or reinforcement learning. This makes it impractical to repeatedly distill from the original large-scale dataset, necessitating incremental KD techniques that integrate new capabilities (e.g., emergent reasoning skills) while preserving previously acquired knowledge. Second, training corpora themselves evolve as datasets grow to include new domains, languages, or specialized content. DD methods must, therefore, adapt to retain rare or foundational data, ensuring linguistic diversity and structural coherence without discarding critical information \citep{jayasuriya2025sparc,kirkpatrick2017overcoming,gururangan2020don,gao2020pile}.

A pressing practical research question is how to develop incremental KD that effectively integrates newly emergent abilities, such as long-form reasoning, measured on benchmarks such as BIG-Bench Lite, into an existing student model without reprocessing the original training set. Important requirements include creating lightweight, sequential knowledge integration strategies. To measure such techniques, a concrete experimental setup involves: (1) distilling a 2024-trained, GPT-4-level teacher model into a smaller student (e.g., 7B parameters), and (2) updating this student exclusively using the teacher’s 2025-acquired data. Retention is then measured on 2024-specific knowledge benchmarks (e.g., MMLU), while adaptation is assessed on 2025-specific capability benchmarks (e.g., new BIG-Bench Lite tasks). 
On the DD side, methods must retain low-resource or time-stamped data that will never re-enter future crawls. Promising approaches for balancing historical and novel knowledge include metadata-aware sampling \citep{he2024seekr,liu2024evolving,blowe2024semantic}.
Ultimately, a unified framework is needed to jointly optimize student model updates and data curation strategies. This would allow for the robust estimation of retained knowledge based on controlled experiments as well as ensure distilled models scale efficiently with emerging LLMs and adaptive data environments.


\paragraph{Non-Traditional Knowledge Distillation Frameworks.}
Transitioning beyond static one-teacher, one-student paradigms introduces challenges in stability, scalability, and knowledge coherence, while also opening new research directions. Simultaneous teacher-student training may lead to unstable convergence due to conflicting objectives, motivating the design of tailored loss functions or staged curricula \citep{li2023unlock,li2024bild,zhang2018deep}. A practical question is how to structure joint training protocols that promote synergy rather than interference, such as gradually exposing the student to different teacher knowledge. In self-distillation, the risk of amplifying biases from uncalibrated self-generated labels raises the need to assess mitigation techniques, including filtering low-confidence predictions and tracking fairness metrics over time \citep{arazo2020pseudo,allen2020towards}.

Multi-teacher frameworks pose distinct challenges such as managing conflicting outputs from domain-specialized models. A key question is how to evaluate and preserve reasoning fidelity in the student across diverse tasks. For instance, reasoning benchmarks like HotpotQA \citep{yang2018hotpotqa} or GSM8K \citep{cobbe2021gsm8k} can reveal whether the student retains step-by-step reasoning abilities. Partial teacher availability due to privacy or deployment constraints also motivates exploration of sequential or asynchronous distillation setups. Computational overhead further increases with teacher count, encouraging research into sparse teacher selection. Finally, tracing individual teacher contributions remains difficult. Approaches such as influence-function analysis or leave-one-teacher-out ablations could help identify knowledge origins, building on early efforts like \citet{wadhwa2025taught}. Addressing these challenges with interpretable and efficient methods will help make non-traditional KD frameworks both robust and scalable.

\paragraph{Architectural Mismatch.}
A central hurdle arises from the difference in architectures between teacher and student models. Expansive Transformer-based LLMs, often optimized for long-context modeling, have distinct inductive biases that smaller or alternative architectures (e.g., non-autoregressive models or mixture-of-experts) may not share \citep{tay2022efficient,raiaan2024review}. Traditional KD methods can struggle under these conditions, as the student model may lack the structural capacity or design features to fully absorb the teacher’s capabilities \citep{fedus2022switch,gu2017non}. This challenge extends to DD, where synthetic or compressed data produced for a specific LLM architecture may fail to transfer effectively to models with different configurations. Attempting to create a single distilled dataset for multiple downstream purposes, from general pretraining to specialized instruction-tuning, only amplifies this problem \citep{sucholutsky2021soft,li2023synthetic}.

Moving forward, three research questions can steer progress toward architecture-aware distillation. First, what alignment metrics best quantify fidelity of transferred reasoning steps when teacher and student architectures diverge, and how can they be integrated into distillation objectives? Second, what loss functions can be re-weighted adaptively to consider a student’s shifting capacity, such that token-level imitation is not prioritized more than transferring higher-level patterns? Third, what principles should guide the generation or selection of distilled datasets so that their linguistic and structural properties remain informative across heterogeneous model families? Addressing these questions will help close the performance gap between heterogeneous teachers and students while preserving the efficiency gains that motivate distillation in the first place.


\paragraph{Knowledge Distillation and Foundation Agents}
Knowledge distillation is a promising approach for overcoming key deployment challenges faced by LLM-based agent systems. Agents designed for tasks such as reasoning, planning, tool manipulation, and interaction with complex environments typically require substantial computational resources~\citep{liu2025advances,li2025review}. By transferring these sophisticated capabilities from large teacher models into more compact student models, knowledge distillation facilitates the development of efficient yet effective agents suitable for deployment on resource-constrained hardware. This method is especially valuable for embodied agents operating in real-time contexts, where minimizing latency directly impacts performance~\citep{liu2024benchmark}, or when deploying agents on edge devices with limited computational capacities.

A crucial consideration during distillation is selectively preserving essential elements such as planning pathways, reasoning chains, and tool-usage patterns, which collectively enable agents to perform effectively in complex tasks~\citep{liu2025advances}. Additionally, multi-agent systems may benefit from targeted distillation strategies that integrate specialized knowledge from multiple teacher agents~\citep{chen2024magdi} into a unified and efficient student model capable of versatile operations. However, one notable technical challenge is ensuring the distilled agent maintains the nuanced interplay between decision-making processes and real-world actions—elements that underpin the emergent capabilities characteristic of high-performing autonomous agents. Future research could explore hybrid distillation methods, combining traditional knowledge distillation techniques with reinforcement learning fine-tuning, thus ensuring robustness and adaptability of the distilled agents in dynamic environments.

\paragraph{Evaluation Gaps.}
Existing evaluation frameworks for distillation often fail to capture the diverse capabilities expected of modern LLMs. For KD, existing benchmarks prioritize narrow metrics, such as task accuracy, while overlooking critical dimensions like coherence, ethical alignment, and zero-shot generalization \citep{chang2024survey, liu2023trustworthy}. In DD, common metrics like token overlap or perplexity fail to evaluate whether distilled datasets preserve deeper reasoning abilities, such as chain-of-thought logic or emergent behaviors inherent to large-scale pertaining \citep{qin2025scaling}. 
Additionally, current distilled model evaluations lack systematic bias and fairness assessments across groups. Standard benchmarks measure only aggregate performance, ignoring whether distillation maintains equitable outcomes for underrepresented populations. This oversight is critical since compression may disproportionately impair performance on minority cases and sensitive attributes. 

Addressing these gaps requires developing holistic evaluation protocols that measure nuanced LLM capabilities, including contextual reasoning, adaptability to unseen tasks, and alignment with ethical guidelines, ensuring distilled models and datasets retain the sophistication of their source systems \citep{dong2024safeguarding,chang2024survey}. 
It is also important to incorporate evaluation frameworks with fairness-specific metrics alongside traditional performance assessments.

\paragraph{Practical Barriers to Adoption. } 

While knowledge distillation and data distillation have shown impressive empirical results, their widespread adoption remains limited, particularly among academic researchers and practitioners with constrained resources, mainly due to the following reasons. 

\noindent\textbf{Computational Requirements.} Many recent KD/DD methods, especially those involving large LLMs, are developed using substantial computational infrastructure, often involving multi-GPU or TPU clusters and extended training times. For example, DistilBERT required 8 V100 GPUs running continuously for 90 hours to distill knowledge from BERT-base \citep{sanh2019distilbert}. Vicuna training costs around \$300 for the 13B model \citep{chiang2023vicuna}, demonstrating the financial investment required even for moderately-sized models. More efficient single-GPU approaches like T5 distillation can compress models in 16 hours on a single A100 \citep{nawrot-2023-nanot5}, but still represent significant computational investments for many research groups.  Historical examples like GPT-2 distillation, with training costs of \$256 per hour, demonstrate the substantial financial barriers to reproducing state-of-the-art distillation methods. 
Additionally, extracting knowledge from teacher LLMs through paid APIs is costly. Building a comprehensive distilled dataset can require thousands of queries, and - depending on prompt complexity, call volume, and dataset size - the total bill can quickly rise into the thousands or even tens of thousands of dollars.
In Table~\ref{tab:kd-computational-requirements}, we list some examples of KD of current LLMs, illustrating the substantial computational and financial requirements across different approaches. These requirements pose significant barriers for smaller research groups or institutions.

\noindent\textbf{Code and Data Availability.} Although a few works, such as DistilBERT, provide publicly available implementations and pretrained models, many others do not release code, model checkpoints, or distilled datasets. This lack of transparency hinders reproducibility, limits benchmarking, and increases the burden of implementation. 

\noindent\textbf{Reproducibility Challenges.} Even when code is available, differences in seemingly minor aspects, such as initialization schemes, optimizer settings, or data preprocessing, can lead to large variations in performance. This is particularly true for DD methods, where the sensitivity to hyperparameters and training schedules is often underreported.

\begin{table}[!ht]
\centering
\caption{Hardware requirements, training times for distilled models across different LLM scales.}
\footnotesize
\renewcommand{\arraystretch}{1.1}
\begin{tabular}{p{1.8cm}p{2.4cm}p{3.3cm}p{2.7cm}p{1.8cm}}
\toprule
\textbf{Method} & \textbf{Teacher $\rightarrow$ Student} & \textbf{Dataset \& Technique} & \textbf{Hardware Requirements} & \textbf{Training Time / Cost} \\ 
\midrule
\textbf{DistilBERT} &
BERT-base (110M) $\rightarrow$ DistilBERT (66M) &
BookCorpus + English Wikipedia
\newline $\bullet$ Classical knowledge distillation
\newline $\bullet$ Soft-logit loss + MLM loss &
8$\times$ V100 16GB GPUs
\newline Very large batch sizes (4K examples) &
90 hours
\newline (3.75 days) \\ 
\midrule
\textbf{T5 Distillation} &
T5-Large (770M) $\rightarrow$ T5-Small (60M) &
C4 dataset (text-to-text tasks)
\newline $\bullet$ Sequence-to-sequence distillation
\newline $\bullet$ Multi-task learning &
1$\times$ A100 GPU
\newline (efficient single-GPU setup) &
16 hours
\newline (0.67 days) \\ 
\midrule
\textbf{Alpaca-LoRA} &
LLaMA-7B $\rightarrow$ 6-layer student &
Stanford Alpaca + Evol-Instruct (52K)
\newline $\bullet$ LoRA/QLoRA (rank-8, 4-bit)
\newline $\bullet$ Parameter-efficient fine-tuning &
1$\times$ RTX 4090 24GB
\newline ($\leq$8 GB GPU RAM usage) &
6 hours
\newline (single GPU) \\ 
\midrule
\textbf{Vicuna} &
LLaMA-13B $\rightarrow$ Vicuna-13B &
ShareGPT conversations (70K)
\newline $\bullet$ Supervised fine-tuning
\newline $\bullet$ Multi-turn dialogue training &
4$\times$ A100 40GB GPUs
\newline Context length expanded to 2048 &
2 days estimated \\ 
\midrule
\textbf{GPT-4 Distillation} &
GPT-4 (1.76T est.) $\rightarrow$ Research models &
API-based distillation
\newline $\bullet$ Reasoning model replication
\newline $\bullet$ Fast iteration cycles &
Standard GPUs &
\$450, 19 hrs \\ 
\bottomrule
\end{tabular}
\label{tab:kd-computational-requirements}
\end{table}

\section{Conclusion}\label{sec:conclusion}

In this survey, we examined the intersection of Knowledge Distillation (KD) and Dataset Distillation (DD) for large language models (LLMs), demonstrating how these complementary approaches address critical challenges in computational efficiency, data scalability, and retention of advanced capabilities. Through an analysis of methodological innovations across both fields, including multi-teacher architectures, rationale-based guidance, adaptive teacher-student co-evolution, and advanced data filtering or synthesis techniques, we present a comprehensive framework for compressing model size and training data while preserving essential LLM strengths, such as contextual reasoning, cross-domain generalization, and linguistic diversity.

Despite rapid progress, several open challenges remain. Preserving emergent abilities in smaller models calls for more sophisticated approaches to capture the nuanced distributions and structural dependencies present in teacher networks. Similarly, generating or filtering distilled data to balance coverage of rare phenomena with overall efficiency demands hybrid or adaptive strategies that continuously align with evolving teacher models and tasks. Ethical and trustworthiness considerations, including mitigating biases and preventing the inheritance of harmful hallucinations, further underscore the importance of carefully calibrated distillation protocols and uncertainty-aware techniques.

Looking ahead, addressing these issues will require novel evaluation frameworks that go beyond standard accuracy metrics to assess deeper language understanding, compositional reasoning, and alignment with societal values. As LLMs are deployed in ever more diverse and critical applications, the continued integration of KD and DD, together with robust checks on model safety and domain appropriateness, represents a promising avenue for sustainable, resource-efficient NLP systems. By combining theoretical insights with real-world validations, future research can help ensure that distillation remains an empowering rather than constraining force for next-generation LLMs.


\backmatter









\section*{Declarations}




\bmhead{Funding}
This work was partially supported by the U.S. National Science Foundation (NSF) [DMS-1925066, DMS-1903226, DMS-2124493, DMS-2311297, DMS-2319279, DMS-2318809, DMS-2101104, DMS-2138854], the National Institutes of Health (NIH) [R01GM152814], and the Institute of Education Sciences (IES) [R305C240010].

\bmhead{Author contribution} 
The survey scope and overall structure were defined by LF and XY. All authors contributed to the literature search and manuscript writing. PM, TL, and WZ supervised the project, refined the survey outline, and provided critical revisions.

\bmhead{Conflict of interest}
The authors declare that there is no Conflict of interest.

\bmhead{Ethics approval and consent to participate} 
Not applicable.

\bmhead{Data/Code/Materials availability} 
Not applicable.

\bibliography{sn-bibliography}

\end{document}